\documentclass[aps,,superscriptaddress]{revtex4-2}
\usepackage{amsmath,amssymb,graphicx,bm}
\usepackage{hyperref}
\usepackage{xcolor}
\usepackage{hyperref}
\usepackage{nameref}
\usepackage{booktabs}
\usepackage{subcaption}
\usepackage{float}
\usepackage{caption}
\usepackage{tabularx}
\usepackage{booktabs}
\usepackage{graphicx}
\usepackage{geometry}
\usepackage{caption}
\usepackage{comment}
\usepackage{tikz}
\usepackage[export]{adjustbox}  

\begin{document}
\title{First-Extinction Law for Resampling Processes}

\author{Matteo Benati}
\affiliation{DIAG, Sapienza University of Rome, Italy}

\author{Alessandro Londei}
\affiliation{Sony CSL Rome, CREF, Rome, Italy}

\author{Denise Lanzieri}
\affiliation{Sony CSL Rome, CREF, Rome, Italy}

\author{Vittorio Loreto}
\affiliation{Sony CSL Rome, CREF, Via Panisperna 89/A, Rome, Italy}
\affiliation{Sony CSL Rome, CREF, Rome, Italy}
\affiliation{Complexity Science Hub, Vienna, Austria}

\begin{abstract}
Extinction times in resampling processes are fundamental yet often intractable, as previous formulas scale as $2^M$ with the number of states $M$ present in the initial probability distribution. We solve this by treating multinomial updates as independent square-root diffusions of zero drift, yielding a closed-form law for the first-extinction time.  We prove that the mean coincides exactly with the Wright--Fisher result of Baxter et. al., thereby replacing exponential-cost evaluations with a linear-cost expression, and we validate this result through extensive simulations. Finally, we demonstrate predictive power for model collapse in a simple self-training setup: the onset of collapse coincides with the resampling-driven first-extinction time computed from the model’s initial stationary distribution. These results hint to a unified view of resampling extinction dynamics.
\end{abstract}

\maketitle
Resampling processes are present in a wide range of natural and artificial systems. When the number of samples is finite, noise accumulates and may be amplified from one iteration to the next, eventually driving the first-extinction time — the first step at which a discrete probability distribution loses one of its states. Classical work on allele extinction in population genetics~\citep{ConstableMcKane2015,Meerson2011,whitlock2000fixation,barton2011genetic}  , novelty–richness dynamics~\citep{tria2014dynamics,loreto2016dynamics} and chemical reactions\citep{AssafMeerson2006} has analyzed special cases, and recent discussions of “model collapse” in AI~\citep{zhang2021rethinking,shumailov2024modelcollapse,schaeffer2025positionmodelcollapsedoes,dohmatob2024modelcollapsedemystifiedcase,gerstgrasser2024modelcollapse} highlighted the need for a tractable, quantitative theory that covers large state spaces and diverse initial conditions \citep{HathcockStrogatz2022,HolehouseRedner2024}.
\begin{figure}[t]
    \centering
    \includegraphics[width=.7\linewidth]{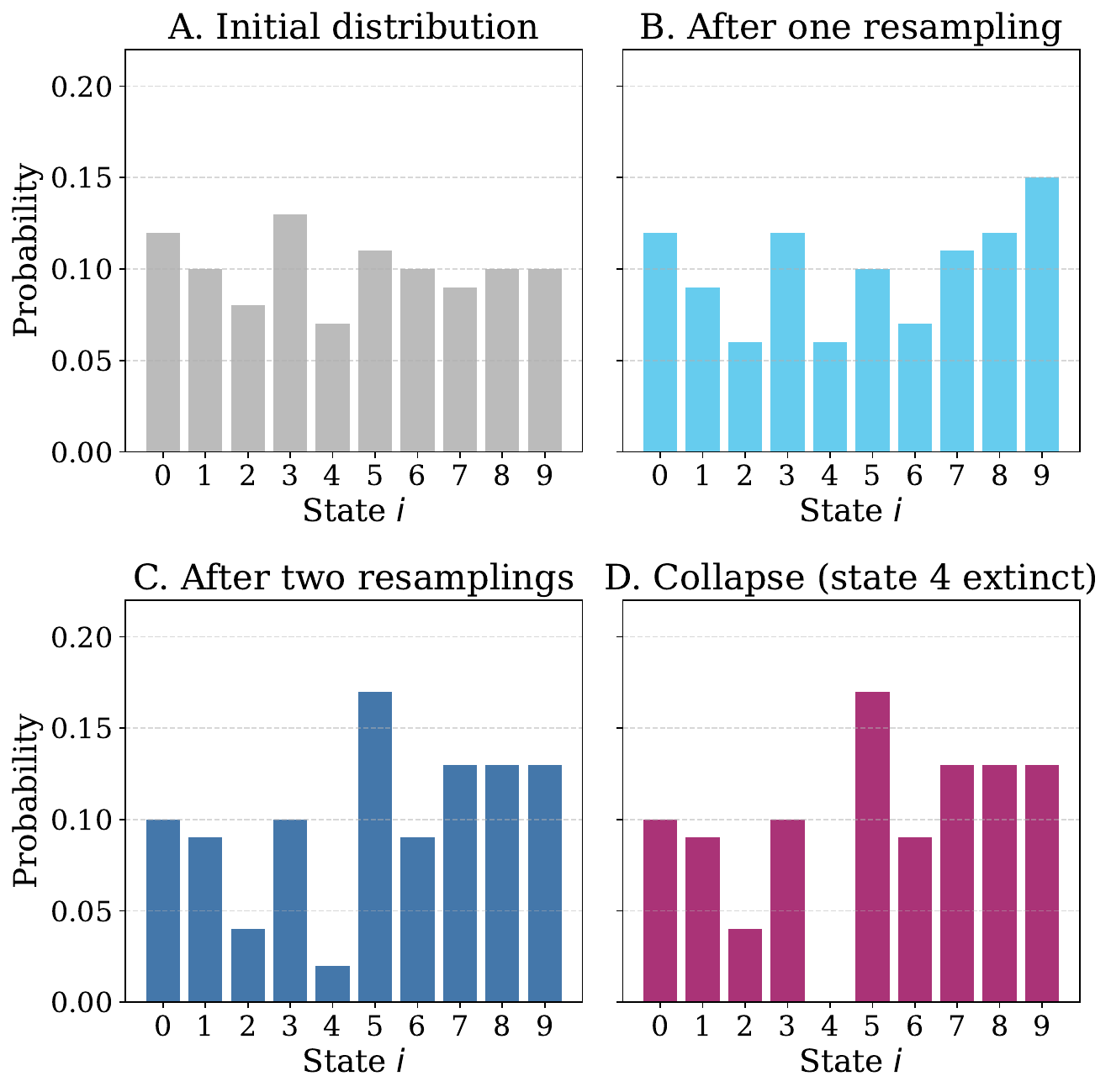}
    \caption{Schematic of the resampling collapse process. Successive sampling steps distort the initial distribution, leading to the extinction of a state at time $\tau=3$.
    }
   \label{fig:resampling_schematic}
\end{figure}
Predicting when a discrete distribution first loses a state under repeated sampling has remained intractable beyond small systems because exact results in population genetics rely on inclusion--exclusion formulas over all state subsets, with computing cost growing as $2^M$~\cite{baxter2007exact,Baxter2008,Tavare1984}. This limits explicit formulas for first-extinction statistics to very small number of states $M$. We resolve this by showing that, under minimal assumptions, we can model correlated Wright--Fisher dynamics through independent zero-drift square-root diffusion ~\citep{fisher_1923,wright_1931,feller1951singular,feller_1952}. From this we derive an exact $\mathcal{O}(M)$ first-extinction time formula that circumvents the computation problems of Baxter's formula. Beyond the technical advance, this result provides a stochastic law for extinction across various fields. The same square-root dynamics are applicable to allele loss in genetics, appears in the Cox–Ingersoll–Ross process in finance, and is the primary cause of disappearance of rare states in AI models during iterative training. We illustrate the breadth of this mechanism by predicting accurately the iteration at which model collapse happens in predictive probabilistic neural networks. Together, these findings identify a general mechanism for rare-event extinction and supply a tractable analytic prediction of first-extinction times in high-dimensional stochastic systems.\\
To model the resampling process, consider a discrete probability distribution \( p(t) \) defined over \( M \) states. At each iteration, \( N \) independent samples are drawn from \( p(t) \), leading to updated probabilities given by
\begin{equation}
    p_i(t+1) = \frac{n_i(t)}{N}
    \label{p(t+1)}
\end{equation}
where $n_i(t)$ is the number of draws for the state i at time t.
This introduces a sampling error whose amplitude is determined by the variance of the binomial distribution governing the process:

\begin{equation}
    \text{Var}(n_i) = N p_i(t) (1 - p_i(t))
\end{equation}

Using Eq.\eqref{p(t+1)}, and the scaling property of variance, \( \text{Var}(aX) = a^2 \text{Var}(X) \), the variance of the reconstructed probability is:

\begin{equation}
    \text{Var}(p_i(t+1)) = \frac{p_i(t)(1 - p_i(t))}{N}
    \label{var}
\end{equation}

We denote the discrete jump in probability as
\[
\Delta p_i = p_i(t+1) - p_i(t)\,.
\]
At each time step \( t \), the probability \( p_i(t) \) is fixed when computing the variance of the update \( \Delta p_i = p_i(t+1) - p_i(t) \). From Eq.\eqref{var}, we have 
\[
\mathbb{E}[\Delta p_i \mid p_i(t)] = 0, \quad \mathrm{Var}[\Delta p_i \mid p_i(t)] = \frac{p_i(t)(1 - p_i(t))}{N}.
\]
The Central Limit Theorem implies that for large \(N\), these small, zero‐mean jumps can be treated as independent Gaussian increments. In the continuous‐time limit
\(\Delta t\to 0\), one thus identifies
\[
\Delta p_i \simeq \sqrt{\frac{p_i(t)(1 - p_i(t))}{N}}\;\Delta W_i(t)\,,
\]
where \(\Delta W_i\) is a Wiener increment with \(\mathbb{E}[(\Delta W_i)^2]=\Delta t\). Most $p_i(0)$ are small, in particular the rarest ones, which are the ones that drive the extinction process. We thus assume $p_i(1 - p_i)\approx p_i$. Taking \(\Delta t\to 0\) yields the SDE
\begin{equation}
\boxed{dp_i(t) = \sqrt{\frac{p_i(t)}{N}} dW_i(t)} 
\label{SDE}
\end{equation}
This is a Feller square-root process with no drift term, more famous as the Cox--Ingersoll--Ross (CIR) model in financial mathematics~\citep{Cox1985,gardiner2009stochastic}. The absence of a drift term in the SDE represents a pure resampling setting, and corresponds to the no-mutation case in genetic drift models~\citep{Ewens2004}. 
Starting from Eq.~\eqref{SDE}, we want to take $M$ different SDEs with starting $p_i(0)$ representing the initial probability distributions. The origin is absorbing in the zero-drift square-root process; we use the Feller first-passage law to 0 to obtain the cumulative distribution function (CDF) for single extinction time~\citep{Masoliver2012,feller1951singular,mazzon2011square}, derived from \ref{SDE}:

\begin{equation}
F_i(\tau) =  1-\gamma\left(1, \frac{2N p_i(t=0)}{ \tau} \right).
\label{eq:CDF}
\end{equation}
Where $\gamma\left(1,x\right)$ is the lower incomplete Gamma function.
Since 

\begin{equation}
\gamma\left(1,x\right) =  1 - e^{-x},
\end{equation}

\eqref{eq:CDF} reduces to:

\begin{equation}
F_i(\tau) =   e^{\frac{-2N p_i(0)}{\tau}}.
\end{equation}
This function represents the probability that state \( i \) is extinct by time \( \tau \). The probability that state \( i \) is not extinct yet by time \( \tau \), $S_i(\tau)$, is given by
\begin{equation}
S_i(\tau) = 1 - F_i(\tau) = 1 - e^{\frac{-2N p_i(0)}{\tau}}.
\end{equation}
Using the independence assumption, the probability that all \( M \) states survive up to time \( \tau \) is given by the product~\citep{schuss2010firstpassage}
\begin{equation}
P(\tau_{\min} > \tau) = \prod_{i=1}^{M} \left( 1 - e^{\frac{-2N p_i(0)}{\tau}} \right).
\label{P_tau_min>tau}
\end{equation}

Then the expected value of the first-extinction time $\tau_{\min}$ is given by

\begin{align}
\langle \tau_{\min} \rangle &= \int_{0}^{\infty} P(\tau_{\min} > \tau) d\tau  \nonumber \\
\end{align}

Substituting:

\begin{equation}
\langle \tau_{\min} \rangle =\boxed{\int_{0}^{\infty} \prod_{i=1}^{M} \left(1 - e^{\frac{-2N p_i(0)}{\tau}} \right) d\tau}\label{theoretical_formula}
\end{equation}

This integral captures the expected time at which the first state reaches zero probability, effectively marking the first-extinction time~\citep{redner2001guide}.\\
\textbf{Scaling in $N$.} Substituting $x=2N/\tau$, Eq.\eqref{theoretical_formula} gives
$\langle \tau_{\min}\rangle = 2N \int_0^\infty \!\left[\prod_{i=1}^M (1-e^{-x p_i(0)})\right] x^{-2}\,dx,$
so $\langle \tau_{\min}\rangle \propto N$ for fixed $\{p_i(0)\}$.

From \eqref{P_tau_min>tau},
we can also derive the CDF for the total first-extinction time $\tau_{min}$ from the survival probability:

\begin{align}
F(\tau_{\min}) 
&= P(\tau_{\min} \leq \tau) \\
&= 1 - P(\tau_{\min} > \tau) \\
&= 1 - \prod_{i=1}^{M} \left(1 - e^{-\frac{2N p_i(0)}{\tau}}\right).
\label{eq:TheoreticalCDF}
\end{align}

\begin{figure}
    \centering
    \includegraphics[width=.7\linewidth]{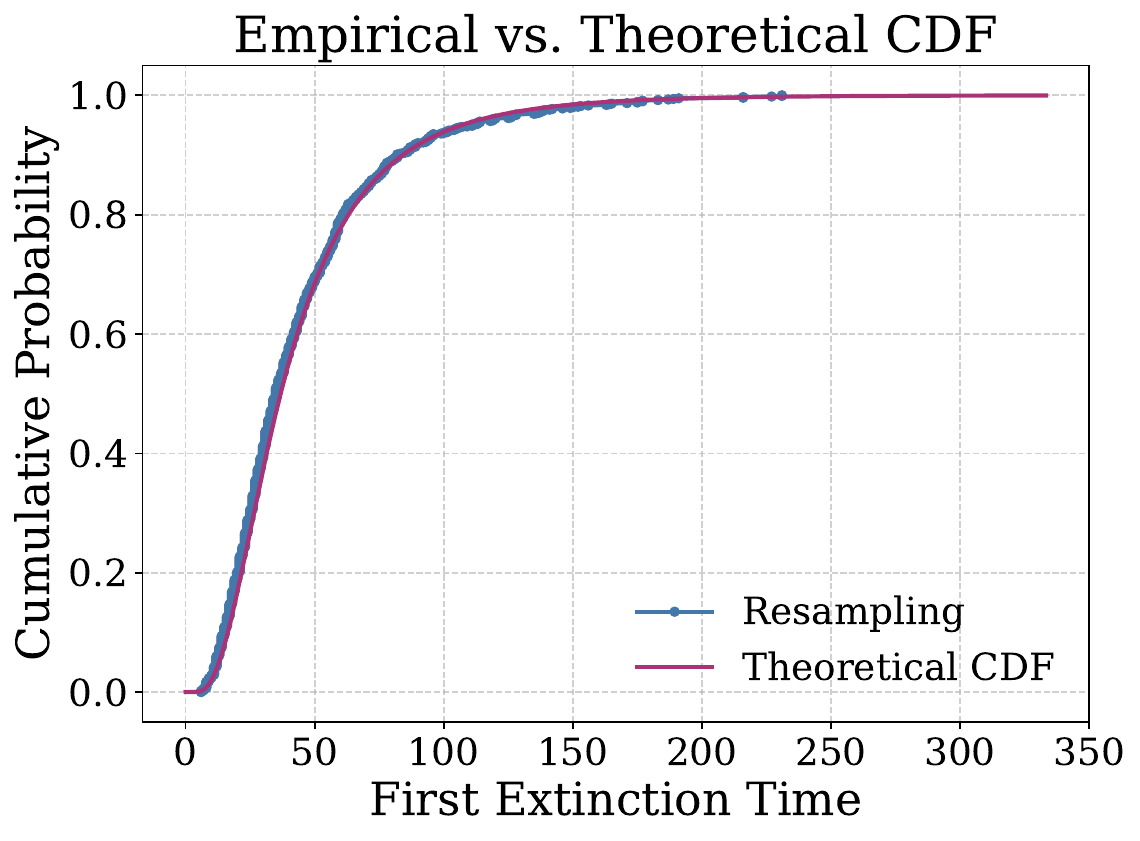}
    \caption{Empirical cumulative distribution of first-extinction times compared to theoretical first-extinction predictions. Theoretical predictions (purple line) closely match the cumulative distributions of resampling (blue dots) first-extinction time, demonstrating strong agreement between theory and experiment. The simulations were conducted with \( M = 100 \) states, \( N = 10^6 \)  samples per iteration, a normalized entropy for the initial probability distribution \( S = 0.90 \) and a total of 1000 simulations.}
    \label{fig:CDF_resampling}
\end{figure}

Our analytical expression for the \emph{mean} first–extinction time (derived under independent square–root diffusions) admits an \(O(M)\) evaluation even for large state spaces, avoiding the naive \(2^{M}\)  inclusion-exclusion expansion present in Baxter et al.~\cite{Baxter2008}. Most importantly, in the Supplementary Material we prove that this mean coincides mathematically with the classical Wright–Fisher result of Baxter for any  $M$ \textbf{(S46-57)}. We also conducted a numerical comparison between the two formulas in many different settings, confirming exact agreement in the $M$ range ($3\leq M \leq 30$), for which the Baxter calculation doesn't require an infeasible computing time. To further test the theory, we simulate resampling collapse using multinomial resampling across a range of initial entropies and system sizes. We then compared these results with the theoretical first-extinction time~\eqref{theoretical_formula}  and CDF~\eqref{eq:TheoreticalCDF}. The simulations shown in Fig. \ref{fig:CDF_resampling} were performed with \( N = 10^6\) samples per iteration and \( M = 100 \) states, using an initial probability distribution \( p_0 \) generated with normalized entropy $S = \frac{H(p(0))}{log(M)} = 0.90$. A total of 1000 independent trials were carried out. To assess the agreement between the different methods, we conducted statistical comparisons of the mean and standard deviation of first-extinction times across resampling and theoretical predictions. Additionally, we performed  the Kolmogorov–Smirnov test to compare the cumulative distribution of first-extinction times obtained from each method. The mean first-extinction times for the resampling process is $ \langle\tau_{resampling}\rangle = 45.62 \pm 1.05$, which is in close agreement with the theoretical prediction $\langle\tau_{min}\rangle=46.40$. The statistical results also show close alignment, with no significant difference between the two cumulative distributions, yielding a KS-statistic $D=0.042$ and a $p_{value} = 0.341$. Fig. \ref{fig:CDF_resampling} is just one of many simulations. We tested our predictions across $(N,M)$ grids and many initial normalized entropy values $S(p_i(0))$ and we find excellent agreement in both mean and CDF shape for $M\ge 5$. We also verified the quantitative limit for small N with small $M$, showing empirically that, the formula breaks, while it yields $p_{value} \geq 0.05$ for $N\geq200$ (See Supplement). These small discrepancies we found for small M and N could be justified by the fact that the theoretical formula follows by a continuum time SDE, while the actual resampling process is a discrete one.

\begin{figure}[h!]
    \centering
    \includegraphics[width=.7\linewidth]{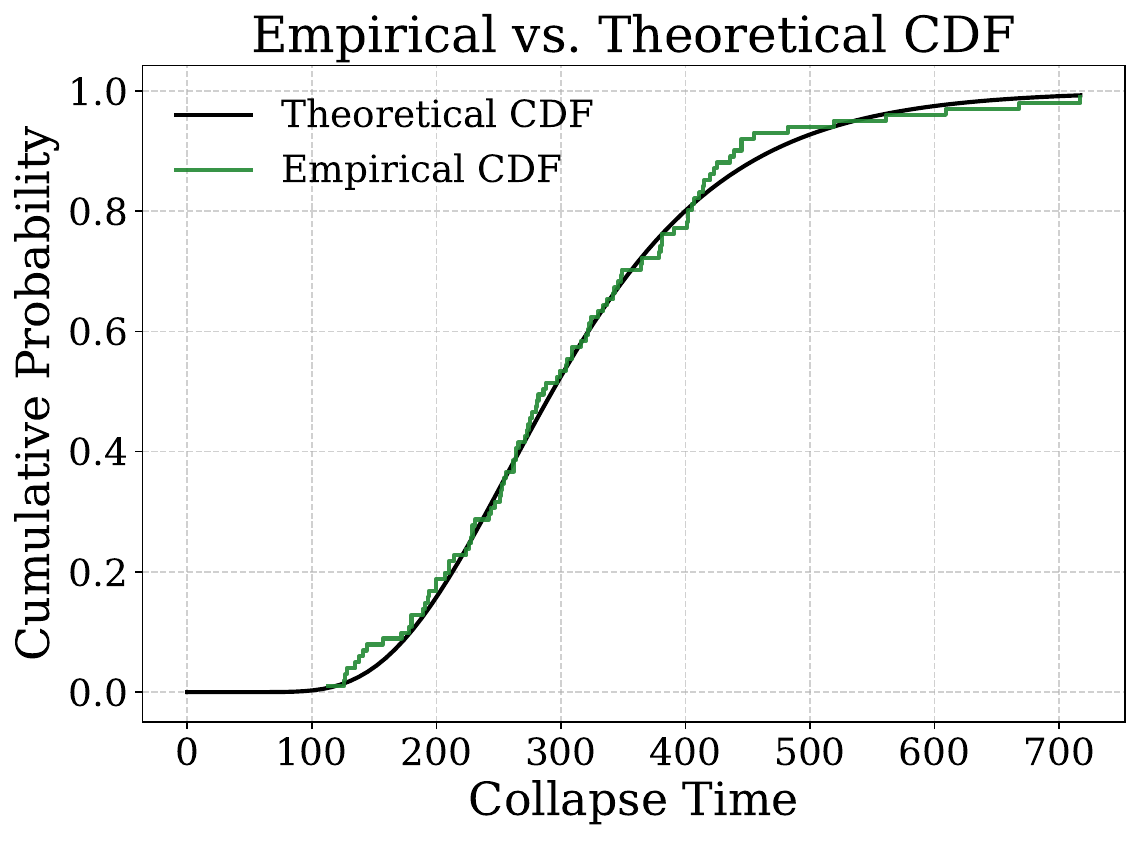}
    \caption{Empirical (green discrete line) versus theoretical (black continuous line) cumulative distribution function (CDF) of the collapse/first-extinction time for the stationary distribution of the Markov chain prediction model. The Markov chain has 30 states, which corresponds to $M=30$. The stationary distribution's normalized entropy is $S=0.85$. The simulation was repeated 100 times, each with a sample size $N = 10^5$ per training cycle. The mean first-extinction time is 310.54 steps, corresponding to a total of $100\cdot 310.54 =31,054$ different training runs.}
\label{fig:markov_collapse}
\end{figure}

\textbf{Model collapse} is a phenomenon that occurs when machine learning models are repeatedly trained on synthetic data (data that is artificially generated). The model starts to lose information because it is learning from its own progressively distorted outputs, rather than from diverse, real-world data. This creates a vicious cycle where the model becomes less and less accurate over time.
Given the novelty of this topic, there is still no consensus on a consistent and rigorous definition of model collapse~\citep{schaeffer2025positionmodelcollapsedoes} and on how to prevent it~\citep{gerstgrasser2024modelcollapse,seddik2024badtrainingsyntheticdata,dohmatob2024strongmodelcollapse,drayson2025machine}. We propose a definition that identifies it with a generalization of the resampling collapse: an irreversible loss of information resulting in a discrepancy between the real data and the model. We will use this definition to test the predictive capabilities of  our first-extinction time formula on this application.  We simulate a network trained to predict sequences generated by a Markov chain with a $30$ by $30$ transition matrix $\mathbf{P}$. The model is first trained on true transitions and subsequently retrained on its predictions. In this case, model collapse occurs when one state probability becomes zero in the Markov chain’s stationary distribution, preventing the model from ever predicting that state again. Due to exponentially increasing computational costs that would be necessary in order to build a large enough statistical pool of model collapse events  —we performed 31,054 training runs— we used a relatively small feed-forward neural network. Architectural details and hyper-parameter schedules are listed in the Supplement.  The training followed cross-entropy minimization with initial stationary distribution entropy of $S = 0.85$ and number of samples $N= 10^5$. We repeated the entire procedure for $100$ independent random seeds. Figure~\ref{fig:markov_collapse} presents the cumulative first-extinction time distribution for the Markov chain's limit probability distribution, comparing empirical results from AI training to the theoretical predictions derived from our model. The theoretical curve takes the initial $p_i(0)$ to be the Markov chain’s stationary distribution. We can see that the empirical cumulative collapse distribution and the theoretical one align extremely closely, with a KS-statistic $D =0.06$ and a p-value $= 0.994$. These results indicate that, in our setting, the first-extinction time is governed primarily by resampling noise, independent of the specific model architecture or its functional error. This finding aligns with the recent analysis of Shumailov  \emph{et al.} \cite{shumailov2024modelcollapse}, who identify finite-sample resampling error as the primary  source of model collapse. Our first-extinction law therefore provides a \textbf{predictor of collapse onset}: it allows us to estimate the precise iteration at which rare events irreversibly vanish, regardless of the model used. While resampling emerges as the primary driver of model collapse, we do not claim it to be the sole cause of it; extending these tests to broader datasets, architectures, and large  language models remains an essential direction for future work. Crucially, by isolating the role of resampling and developing predictive tools for its effects, we lay the foundation for establishing guidelines on the safe and effective use of synthetic data. This step is essential for preventing model collapse and ensuring the long-term reliability and scalability of data-driven AI systems.
We have shown that repeated stochastic resampling induces a collapse of low-probability states in discrete systems, and we have derived an explicit and tractable expression for the distribution of the first-extinction time under minimal assumptions that is usable for many-states systems. Our analytical predictions and numerical simulations confirm that the first-extinction time depends only on the sample size, the number of states, and the initial probability distribution, with no adjustable parameters. We demonstrate that the same square-root stochastic dynamics naturally extends to modern AI models, explaining how finite-sample noise can drive the progressive loss of rare modes in iterative self-training \cite{shumailov2024modelcollapse}. This mechanism provides a compact stochastic description for rare-event extinction, complementing classical results in population genetics~\cite{Tavare1984,Baxter2008},  extinction rates in noise-driven systems~\cite{Meerson2011}, and the dynamics of short-term interest rates in financial models such as the Cox--Ingersoll--Ross process~\cite{Cox1985}. This result clarifies the theoretical foundations of collapse phenomena and motivates the design of collapse-aware algorithms for long-term stability in data-driven generative pipelines. More broadly, the framework quantifies the inevitable loss of diversity in any finite discrete process, offering a compact predictive tool for systems with many states or alleles, from information theory to artificial intelligence and population genetics.
\section*{Acknowledgments}
This work has been supported by PNRR MUR project PE0000013-FAIR.

\bibliographystyle{unsrt}  
\bibliography{main}

\clearpage
\onecolumngrid
\begin{center}
    {\large \bf Supplementary Information for:\\
    First-Extinction Law for Resampling Processes}
\end{center}

\setcounter{section}{0}
\renewcommand{\thesection}{S\arabic{section}}
\setcounter{figure}{0}
\renewcommand{\thefigure}{S\arabic{figure}}

\setcounter{equation}{0}
\setcounter{table}{0}
\setcounter{figure}{0}

\renewcommand{\theequation}{S\arabic{equation}}
\renewcommand{\thetable}{S\arabic{table}}
\renewcommand{\thefigure}{S\arabic{figure}}

\newtheorem{theorem}{Theorem}[section]
\newtheorem{lemma}[theorem]{Lemma}
\newtheorem{definition}[theorem]{Definition}
\newtheorem{remark}[theorem]{Remark}
\newtheorem{prop}{Proposition}

\title{Supplemental Material for\\
\textit{First-Extinction Law for Resampling Processes}}
\author{Matteo Benati}
\affiliation{Your Institution}
\maketitle

\renewcommand{\thesection}{S\arabic{section}}
\renewcommand{\theequation}{S\arabic{equation}}
\renewcommand{\thefigure}{S\arabic{figure}}
\setcounter{section}{0}
\setcounter{equation}{0}
\setcounter{figure}{0}

\section{Analytical Derivation of first-extinction time Distribution}
We want to derive, given a discrete probability distribution, \textbf{the mean time for at least one of the bins to reach zero}. In order to find it we need to have the probability density function of the mean first passage time, and that is what we are going to derive in the next section. For the reasons we argued in the paper we can assume the diffusion term as equal to $\sqrt{\frac{p_i(t)}{N}})$ and model the probability $p_i(t)$ of a given bin evolving over time as the stochastic differential equation (SDE) we used in:
\begin{equation}
    dp_i(t) = \sqrt{\frac{p_i(t)}{N}} dW_i(t),
    \label{SDE1}
\end{equation}
Where $W_i(t)$ is a standard Wiener process and $N$ is the sample size.\\
The first-passage time distribution tells us the probability that $p_i(t)$ reaches zero for the first time $\tau$, starting from a certain initial $p_i(0)$. What we want to derive is the probability distribution of the time necessary, over many different initial $p_i(0)$, to get at least one of them to reach 0.

\subsection{Connection to the Square Root Process}

A general form of the square root process is given by:

\begin{equation}
    dp_i(t) = (bp_i(t) + c)dt + \sqrt{2ap_i(t)} dW_t
\end{equation}

For our process (\ref{SDE1}):
\begin{enumerate}
\item{ $b = c = 0$, since there is no drift term.}
\item{ $a = \frac{1}{2N}$ from the variance structure.}
\end{enumerate}

\subsection{Feller’s calculation for $c\le0$}
For the fundamental solution of the transition density for this process, we will follow Feller's calculation \citep{feller1951singular}. We report a summary of its proof, taking parts also from \citep{mazzon2011square}. For simplicity, we keep the notation as the regular SDE

\begin{equation}
    dx(t) = (bx(t) + c)dt + \sqrt{2ax(t)} dW_t
\end{equation}
with $x(0)$ written as $\bar x$.

Feller considers the parabolic operator
\begin{equation}\label{eq:operator}
L u(t,x):=\frac{\partial^2}{\partial x^2}\bigl(a x\,u(t,x)\bigr)
-\frac{\partial}{\partial x}\bigl((b x+c)\,u(t,x)\bigr)
-\frac{\partial}{\partial t}u(t,x),
\end{equation}
where $x>0$, $t>0$, $a>0$, $b,c\in\mathbb R$.

We give the following definition of a fundamental solution of \eqref{eq:operator}:

\begin{definition}
A \emph{fundamental solution} of \eqref{eq:operator} with pole at $\bar x$ is a function $p(t,x,\bar x)$ such that:
\begin{enumerate}
  \item $p(\cdot,\cdot,\bar x)\in C^2\cap L^1((0,T)\times\mathbb R^+)$ for any $T>0$;
  \item $\partial_t p(t,\cdot,\bar x)\in L^1(\mathbb R^+)$ for any $t>0$;
  \item $p(0,x,\bar x)=\delta_{\bar x}$ in the sense that, for any \(\varphi\in C_b(\mathbb R^+)\),
  \begin{equation}\label{eq:initial}
    \lim_{(t,x)\to(0,\bar x)}\int_{0}^{\infty}p(t,x,\bar x)\,\varphi(y)\,dy=\varphi(\bar x),
  \end{equation}
  and
  \begin{equation}\label{eq:homogeneous}
    L p(t,x,\bar x)=0,
    \quad (t,x)\in(0,\infty)^2.
  \end{equation}
\end{enumerate}
In this case, $p(t,x,\bar x)$ is also said to be a fundamental solution of the PDE
$$
\partial_t u(t,x)=\frac{\partial^2}{\partial x^2}\bigl(a x\,u(t,x)\bigr)
-\frac{\partial}{\partial x}\bigl((b x+c)\,u(t,x)\bigr)
$$
with initial condition \eqref{eq:initial}.
\end{definition}

\subsection{Laplace transform}

Assuming the existence of $p(t,x,\bar x)$, define for $s>0$ its Laplace transform:
\begin{equation}\label{eq:omega}
\omega(t,s,\bar x)=\int_{0}^{\infty}e^{-s x}\,p(t,x,\bar x)\,dx.
\end{equation}
By differentiating under the integral:
$$
\partial_t\omega(t,s,\bar x)=\int_{0}^{\infty}e^{-s x}\partial_t p(t,x,\bar x)\,dx.
$$
Integrating \eqref{eq:homogeneous} over $[y,1]$ in $x$ and taking the limit $y\to0$ introduces the \emph{flux at the origin}
$$
f(t)=-\lim_{x\to0}\Bigl(\partial_x(a x p)+ (b x+c)p\Bigr)(t,x,\bar x).
$$
A detailed calculation then shows
\begin{equation}\label{eq:pde-omega}
\partial_t\omega + s(as-b)\partial_s\omega = -c s\,\omega + f(t),
\end{equation}
with initial condition
\begin{equation}\label{eq:omega-init}
\lim_{t\to0}\omega(t,s,\bar x)=e^{-s\bar x}.
\end{equation}
Feller solves \eqref{eq:pde-omega} by the method of characteristics, obtaining
a general formula (see \cite{feller1951singular}):
\begin{equation}\label{eq:omega-general}
\begin{split}
\omega(t,s,\bar x)=&\Bigl(\tfrac{b}{as(e^{bt}-1)+b}\Bigr)^{c/a}
\exp\bigl\{-\tfrac{s b \bar x\,e^{bt}}{as(e^{bt}-1)+ b}\bigr\}\\
&+\int_0^t f(\tau)\Bigl(\tfrac{b}{as(e^{b(t-\tau)}-1)+b}\Bigr)^{c/a}\,d\tau.
\end{split}
\end{equation}
For $b=0$ one takes the limit $b\to0$ in \eqref{eq:omega-general}.

\subsection{Calculation of $f$ for $c\le0$ (subsection 2.2)}

\begin{theorem}
\label{thm:f-equation}
If $c\le0$, then $\omega(t,s,\bar x)$ in \eqref{eq:omega-general} corresponds to a fundamental solution only if
\begin{equation}\label{eq:f-integral}
\exp\bigl\{-\tfrac{\bar x b}{a(1-e^{-bt})}\bigr\}+\int_0^t f(\tau)\Bigl(\tfrac{e^{bt}-1}{e^{b(t-\tau)}-1}\Bigr)^{c/a}d\tau=0.
\end{equation}
\end{theorem}

[Sketch of proof]
Imposing the vanishing of $\omega$ as $s\to\infty$ yields \eqref{eq:f-integral}.

\begin{theorem}
The integral equation \eqref{eq:f-integral} has the unique solution
\begin{align}\label{eq:f-solution}
&f(t)=-\frac b{\Gamma(1-c/a)}e^{-bt}(1-e^{-bt})^{-1}
\\&\cdot\Bigl(\tfrac{\bar x b}{a(1-e^{-bt})}\Bigr)^{(a-c)/a} 
\exp\Bigl\{-\tfrac{\bar x b}{a(1-e^{-bt})}\Bigr\}.
\end{align}
\end{theorem}

\subsection{Laplace transform with $f$ inserted}

Substituting \eqref{eq:f-solution} into \eqref{eq:omega-general} one obtains
\begin{align}\label{eq:omega1}
\omega_1(t,s,\bar x)=&\Bigl(\tfrac{b}{as(e^{bt}-1)+b}\Bigr)^{c/a}
\exp\Bigl\{-\tfrac{s b \bar x\,e^{bt}}{as(e^{bt}-1)+ b}\Bigr\}
\\&\,\cdot\Gamma\Bigl(1-\tfrac{c}{a};\tfrac{b\bar x e^{bt}}{a(e^{bt}-1)}\tfrac{b}{sa(e^{bt} -1) +b}\Bigr),
\end{align}
where $\Gamma(\nu;z)=\frac1{\Gamma(\nu)}\int_0^z e^{-x}x^{\nu-1}dx$ is the upper incomplete Gamma function.

\subsection{Inversion to obtain $p_1$ (subsection 2.4)}

Inverting \eqref{eq:omega1} via Bessel functions leads to the final form:
\begin{equation}\label{eq:p1-final}
\begin{aligned}
p_1(t,x,\bar x)&=\frac{b}{a(e^{bt}-1)}\exp\Bigl\{-\tfrac{b(x+\bar x e^{bt})}{a(e^{bt}-1)}\Bigr\}
\Bigl(\tfrac{e^{-bt}x}{\bar x}\Bigr)^{\frac{c - a}{2a}}\\
&\times I_{1-c/a}\Bigl(\tfrac{2b}{a(1-e^{-bt})}\sqrt{e^{-bt}x\bar x}\Bigr),
\end{aligned}
\end{equation}
where $I_{\nu}$ is the modified Bessel function of the first kind. Re-writing this in the original terms of our work $x = p_i$ and $\bar x = p_i(0)$:

\begin{equation}
\begin{aligned}
q_1(t, p_i; p_i(0)) = &\frac{b}{a(e^{bt} - 1)} \exp \left( - \frac{b(p_i + p_i(0)e^{bt})}{a(e^{bt} - 1)} \right)
\\\left( \frac{e^{-bt}p_i}{p_i(0)} \right)^{\frac{c - a}{2a}}&\times I_{1 - c/a} \left( \frac{2b}{a(1 - e^{-bt})} \sqrt{e^{-bt}p_i p_i(0)} \right)
    \end{aligned}
\end{equation}
and integrating this we obtain 
\begin{equation}
\int_{0}^{\infty} q_{1}(t, p; p_i(0)) \, dp_i(0)
= \gamma\!\Bigl(1 - \frac{c}{a} ; \frac{b \,p_i(0)\, e^{-b t}}{a\bigl(e^{b t} - 1\bigr)}\Bigr)
\label{eq:1.8}
\end{equation}
\\
Where $\gamma(s, x)$ is the \textbf{lower incomplete Gamma function}:
\begin{equation}
    \gamma(s, x) = \int_0^x t^{s-1} e^{-t} dt
\end{equation}
This fundamental solution is a \emph{defective density}, in the sense that this is does not integrate to $1$ (defective density). In fact we are missing one key part of this, which is the probability of reaching exactly $0$. We can correct this by saying that the probability of ending at 0 at a certain point has a pdf $f_i(\tau)$ that normalizes $q_1$, which means its a Dirac delta on $0$ with the weight equal to its CDF:
\begin{equation}
    F_i(\tau) = 1 - \gamma\!\Bigl(1 - \frac{c}{a} ; \frac{b \,p_i(0)\, e^{b t}}{a\bigl(e^{b t} - 1\bigr)}\Bigr)
\end{equation}
This is the result obtained by Feller\citep{feller1951singular,mazzon2011square}.
\subsection{Simplifications with $c = 0$ and $b \to 0$}
Returning to our original SDE
\begin{equation}
    dp_i(t) = \sqrt{\frac{p_i(t)}{N}} dW_i(t),
    \label{SDE2}
\end{equation}

We have $c = 0$, $b = 0$ and $a = \frac{1}{2N}$. Substituting $c$, $F_i(\tau)$ simplifies to:

\begin{equation}
    F_i(\tau) = 1 - \gamma\Bigl(1, \frac{b p_i(0) e^{b\tau}}{a(e^{b\tau} - 1)}\Bigr)
\end{equation}

In our process b is also 0, but we have to be more careful with it, since we need to take the limit $b \to 0$. We use the expansion:

\begin{equation}
    \lim_{b \to 0} \frac{b}{e^{b\tau} - 1} = \frac{1}{\tau},
\end{equation}

which simplifies the argument of the lower incomplete Gamma function:

\begin{equation}
    \lim_{b \to 0} \frac{b p_i(0) e^{b\tau}}{a(e^{b\tau} - 1)} = \frac{p_i(0)}{a \tau}
\end{equation}

Thus, the collapse CDF becomes:

\begin{equation}
    F_i(\tau) = 1-\gamma\left(1, \frac{p_i(0)}{a \tau} \right)
\end{equation}

Since $\gamma(1, z) = 1 - e^{-z}$ and $a = \frac{1}{2N}$, we obtain:

\begin{equation}
    F(\tau) = e^{\frac{-2N p_i(0)}{\tau}}
\end{equation}

Taking the derivative with respect to $\tau$, we obtain the \textbf{first-extinction time probability density function:}

\begin{equation}
    f(\tau) = \frac{1}{\Gamma(1)} \left( \frac{2N p_i(0)}{\tau^2} e^{- \frac{2N p_i(0)}{\tau}} \right)
\end{equation}

Since $\Gamma(1) = 1$, we obtain:

\begin{equation}
    f(\tau) = \frac{2N p_i(0)}{\tau^2} e^{- \frac{2N p_i(0)}{\tau}}
\end{equation}
This PDF accounts for just one single initial $p_i(0)$, now we want to go further by using all PDFs for each initial condition.

\subsection{PDF and CDF for collapse}
Since we want to find the average minimum first passage time for M different initial $p_i(0)$ probabilities we need to find the probability density function of this minimum. In order to do that we use the survival function and the fact that the M different processes are assumed as independent. As we already discussed, the cumulative distribution function (CDF) is given by:
\begin{equation}
    F_i(\tau) =  1-\gamma\left(1, \frac{p_i(0)}{a \tau} \right)
\end{equation}
The CDF of $\tau_i$ and its complement (survival function) are:
\begin{equation}
    F_i(t) = P(\tau_i \leq t), \quad 1 - F_i(t) = P(\tau_i > t)
\end{equation}
where $\tau_i$ represents a first-extinction time sample.
For a set of first-extinction time samples, the probability that all elements exceed $t$ is:
\begin{equation}
    \prod_i P(\tau_i > t) = P(\{\tau_i\} > t) = P(\tau_{\min} > t)
\end{equation}
This implies:
\begin{equation}
    1 - P(\tau_{\min} > t) = P(\tau_{\min} \leq t) = F_{\min}(t)
\end{equation}

The probability density function (PDF) of the minimum value is obtained via differentiation:
\begin{equation}
    \frac{d}{dt} F_{\min}(t) = f_{\min}(t)
\end{equation}

The expected minimum value is given by:
\begin{equation}
    \langle \tau_{\min} \rangle = \int_0^{\infty} t f_{\min}(t) dt
\end{equation}
We choose:
\begin{equation}
    u = t, \quad dv = f_{\min}(t) dt
\end{equation}
which gives:
\begin{equation}
    du = dt, \quad v = \int f_{\min}(t) dt = F_{\min}(t)
\end{equation}
Applying integration by parts:
\begin{equation}
    \int_0^{\infty} t f_{\min}(t) dt = \left[ t F_{\min}(t) \right]_0^{\infty} - \int_0^{\infty} F_{\min}(t) dt
\end{equation}
We integrate the second term:
\begin{equation}
    \int_0^{\infty} t f_{\min}(t) dt =  \left[ t F_{\min}(t) \right]_0^{\infty} - \int_0^{\infty} (1 - P(\tau_{\min} > t)) dt
\end{equation}
Which gives:
\begin{equation}
    \int_0^{\infty} t f_{\min}(t) dt =  \left[ t \right]_0^{\infty} - \left[ t \right]_0^{\infty} +\int_0^{\infty} P(\tau_{\min} > t) dt
\end{equation}
Finally, substituting the survival function:
\begin{equation}
    \langle \tau_{\min} \rangle = \int_0^{\infty} \prod_{i=1}^{M} (1 - F_i(t)) dt
\end{equation}

And substituting the cumulative distribution function:
\begin{equation}
    \langle \tau_{\min} \rangle = \int_0^{\infty} \prod_{i=1}^{M} \\\gamma\left(1, \frac{p_i(0)}{a \tau} \right) dt
\end{equation}

Simplifying and substituting the explicit formula for Gamma:
\begin{equation}
    \langle \tau_{\min} \rangle = \int_0^{\infty} \prod_{i=1}^{M} \left( 1-e^{\frac{-2N p_i(0)}{\tau}}\right) dt
    \label{eq:final_tau}
\end{equation}

This final integral expresses the expected minimum value in terms of the individual survival functions of each sample.

\subsection{Linear scaling of N}
We model each probability \( p_i(t) \) as an independent square-root diffusion:
\begin{equation}
dp_i(t) = \sqrt{\frac{p_i(t)}{N}} \, dW_i(t)
\end{equation}
This implies that the stochastic fluctuations scale as \( 1/\sqrt{N} \), so the typical time to extinction scales as \( N \). More precisely, the mean time to first extinction is:
\begin{equation}
\langle \tau_{\min} \rangle = \int_0^\infty \prod_{i=1}^M \left(1 - e^{-\frac{2N p_i(0)}{\tau}} \right) d\tau
\end{equation}
Using the change of variable \( x = \frac{2N}{\tau} \), the integral becomes:
\begin{equation}
\langle \tau_{\min} \rangle = 2N \int_0^\infty \prod_{i=1}^M \left(1 - e^{-x p_i(0)} \right) \frac{1}{x^2} dx
\end{equation}
which shows that \( \langle \tau_{\min} \rangle \propto N \).
\subsection{Analytical equivalence between Baxter and our formula}
\begin{equation}
\textbf{Claim.}\quad
I_M:=\int_{0}^{\infty}\prod_{i=1}^{M}\Bigl(1-e^{-a_i/\tau}\Bigr)\,d\tau
=\sum_{\varnothing\neq S\subseteq[M]}(-1)^{|S|}
\Bigl(\sum_{i\in S}a_i\Bigr)\ln\!\Bigl(\sum_{i\in S}a_i\Bigr),
\quad a_i>0.
\end{equation}
This is the exact result derived by Baxter et al.~\cite{Baxter2008} we want to prove is equal to ours.

\begin{equation}
x=\frac{1}{\tau},\qquad
f(x):=\prod_{i=1}^{M}\bigl(1-e^{-a_i x}\bigr),\qquad
I_M=\int_{0}^{\infty}\frac{f(x)}{x^{2}}\,dx.
\end{equation}
Change of variables; definition of $f. \textit{As }x\downarrow 0,\ f(x)=(\prod_i a_i)x^M+O(x^{M+1}); \textit{as }x\to\infty,\ f(x)\to1.$ 

\begin{equation}
I_M=\Bigl[-\frac{f(x)}{x}\Bigr]_{0}^{\infty}+\int_{0}^{\infty}\frac{f'(x)}{x}\,dx
=\int_{0}^{\infty}\frac{f'(x)}{x}\,dx.
\end{equation}
One integration by parts; the boundary term vanishes since $f(x)=O(x^M)\textit{ at }0\textit{ and }\frac{f(x)}{x}\to0\textit{ at }\infty\textit{.}$

\begin{equation}
f'(x)=\sum_{j=1}^{M} a_j e^{-a_j x}\prod_{i\ne j}\bigl(1-e^{-a_i x}\bigr).
\end{equation}
Differentiating the product (each summand drops the $j$-th factor).

\begin{equation}
\prod_{i\ne j}(1-e^{-a_i x})
=\sum_{T\subseteq [M]\setminus\{j\}}(-1)^{|T|}\,e^{-\bigl(\sum_{i\in T}a_i\bigr)\,x}.
\end{equation}
Inclusion–exclusion expansion over $U_j=[M]\setminus\{j\}.$

\begin{equation}
f'(x)=\sum_{j=1}^{M} a_j\sum_{T\subseteq U_j}(-1)^{|T|}
\,e^{-(a_j+\sum_{i\in T}a_i)\,x}.
\end{equation}
Insert the expansion into $f'(x)$.This is a finite sum of exponentials.

\begin{equation}
\sum_{T\subseteq U_j}(-1)^{|T|}e^{-(a_j+\sum_{i\in T}a_i)x}
=\sum_{T\subseteq U_j}(-1)^{|T|}
\bigl(e^{-(a_j+\sum_{i\in T}a_i)x}-e^{-a_j x}\bigr).
\end{equation}
We regrouped using$(\sum_{T\subseteq U_j}(-1)^{|T|}=0)$(binomial theorem). Substituting it in $I_M$:

\begin{equation}
I_M=\sum_{j=1}^{M} a_j\sum_{T\subseteq U_j}(-1)^{|T|}
\int_{0}^{\infty}\frac{e^{-(a_j+\sum_{i\in T}a_i)x}-e^{-a_j x}}{x}\,dx.
\end{equation}
Bring the finite sums outside the integral the integral. Each bracket is a Frullani integral.

\begin{equation}
\int_{0}^{\infty}\frac{e^{-\beta x}-e^{-\alpha x}}{x}\,dx=\ln\frac{\alpha}{\beta}\qquad(\alpha,\beta>0).
\end{equation}
Frullani’s integral; used termwise with $(\alpha=a_j,\ \beta=a_j+\sum_{i\in T}a_i)$.

\begin{equation}
I_M=\sum_{j=1}^{M} a_j\sum_{T\subseteq U_j}(-1)^{|T|}
\ln\frac{a_j}{\,a_j+\sum_{i\in T}a_i\,}
=-\sum_{j=1}^{M} a_j\sum_{T\subseteq U_j}(-1)^{|T|}\!
\ln\!\Bigl(a_j+\sum_{i\in T}a_i\Bigr).
\end{equation}
Split the log; the $(\ln a_j)$ part cancels because $(\sum_{T\subseteq U_j}(-1)^{|T|}=0)$.

\begin{equation}
\text{Let }S:=T\cup\{j\}\ (\neq\varnothing)\ \Rightarrow\ 
a_j+\textstyle\sum_{i\in T}a_i=\sum_{i\in S}a_i=:A_S,\quad |S|=|T|+1.
\end{equation}
Reindex the double sum to a single subset sum over nonempty $S\subseteq[M]\textit{.}$

\begin{equation}
I_M=\sum_{\varnothing\neq S\subseteq[M]}(-1)^{|S|}
\Bigl(\sum_{i\in S}a_i\Bigr)\ln\!\Bigl(\sum_{i\in S}a_i\Bigr),
\end{equation}
Which is the claimed formula.

\section{Additional Numerical Analysis}

\subsection{Comparison with Baxter First-Extinction Times}

We compare the mean time to the first extinction event in a discrete stochastic resampling process for different values of the number of states \( M \). Specifically, we contrast two approaches:

\begin{itemize}
  \item The exact result derived by Baxter et al.~\cite{Baxter2008}, based on the multi-allelic Wright–Fisher diffusion, which includes full correlations and enforces the conservation constraint \( \sum_{i=1}^M p_i = 1 \).
  \item Our theoretical formula, based on independent square-root (Feller) diffusion for each \( p_i(t) \).
\end{itemize}

Both models assume neutral dynamics and no mutation. We used many initial conditions, starting with  a flat distribution \( p_i(0) = 1/M \) for all \( i \) (as the initial distribution used in the original paper of Baxter et al. ~\cite{Baxter2008}) and went to increasingly lower initial normalized entropies $S(p(0))$. For comparison with Baxter et al. We report exact mean times in the conventional units of $2N$ (their notation), while our main-text SDE and all simulations measure $\tau$ in \emph{iterations}. When comparing values, we convert units explicitly. Table~\ref{tab:comparison} reports the mean extinction times obtained using both methods for selected values of \( M \). The theoretical value is computed using the one-dimensional integral in \ref{eq:final_tau}, which is numerically stable and computationally efficient even for large \( M \). As shown, the two methods yield identical results up to numerical precision. This confirms that the square-root SDE formula is quantitatively correct at small values of \( M \) and $N$ when evaluating the mean first-extinction time. We extended the analysis to non-flat distribution and they yielded perfect agreement too for Tab.\ref{tab:comparison2}, Tab.\ref{tab:comparison3} and Tab.\ref{tab:comparison4}. This outstanding accuracy confirms that our assumptions are reasonable and yield the same results as the exact Baxter formula. This suggests that the small discrepancies we notice for small $N$ could be due to differences between the continuous approximation and the discrete one, where we have discrete jumps for the probabilities, rather than continuous ones.
\begin{table}[h!]
  \centering
  \caption{Comparison between the Baxter exact formula and the resampling-based formula for the mean time to first extinction for the flat initial distribution. Since the extinction time is linear in $2N$,the mean times to first extinction are indicated as multiples of $2N$. }
  \label{tab:comparison}
  \begin{tabular}{c|c|c}
    \toprule
    \( M \) & Baxter formula & Resampling formula \\
    \midrule
    3  & 0.5753 & 0.5753 \\
    4& 0.3397& 0.3397\\
 5& 0.2333&0.2333\\
    10 & 0.0810& 0.0810\\
    20 & 0.0311& 0.0311\\
    25 & 0.0231& 0.0231\\
    \bottomrule
 30& 0.0182&0.0182\\
  \end{tabular}
\end{table}

\begin{table}[h!]
  \centering
  \caption{Comparison between the Baxter exact formula and the resampling-based formula for the mean time to first extinction for initial distributions of normalized entropy $S(p_i(0)) = 0.90$ of increasing M. Since the extinction time is linear in $2N$,the mean times to first extinction are indicated as multiples of $2N$. }
  \label{tab:comparison2}
  \begin{tabular}{c|c|c}
    \toprule
    \( M \) & Baxter formula & Resampling formula \\
    \midrule
    3  & 0.4817 & 0.4817 \\
    4& 0.2577& 0.2577\\
 5& 0.1571&0.1571\\
    10 & 0.0238& 0.0238\\
    15 & 0.0065& 0.0065\\
  \end{tabular}
\end{table}

\begin{table}[h!]
  \centering
  \caption{Comparison between the Baxter exact formula and the resampling-based formula for the mean time to first extinction for initial distribution of normalized entropy $S(p_i(0)) = 0.80$ of increasing M. Since the extinction time is linear in $2N$,the mean times to first extinction are indicated as multiples of $2N$. }
  \label{tab:comparison3}
  \begin{tabular}{c|c|c}
    \toprule
    \( M \) & Baxter formula & Resampling formula \\
    \midrule
    3  & 0.3837 & 0.3837 \\
    4& 0.1693 & 0.1693\\
 5& 0.00209&0.00209\\
    10 & 0.00016& 0.00016\\
    15 & 0.000127& 0.000127\\
  \end{tabular}
\end{table}

\begin{table}[h!]
  \centering
  \caption{Comparison between the Baxter exact formula and the resampling-based formula for the mean time to first extinction for initial distribution of normalized entropy $S(p_i(0)) = 0.70$ of increasing M. Since the extinction time is linear in $2N$,the mean times to first extinction are indicated as multiples of $2N$.}
  \label{tab:comparison4}
  \begin{tabular}{c|c|c}
    \toprule
    \( M \) & Baxter formula & Resampling formula \\
    \midrule
    3  & 0.1083 & 0.1083 \\
    4& 0.1084 & 0.1084\\
 5& 0.0757& 0.0757\\
    10 & 0.000217 & 0.000217\\
    15 & 0.0000589 & 0.0000589\\
  \end{tabular}
\end{table}
It is important to notice that, differently from the flat case, the initial distribution of Tables II to IV are selected just for entropy and thus their mean first-extinction time is specific to the initial distribution drawn at the beginning of each resampling process. For this reason it can happen that the first-extinction time does not scale in the same way, since each mean time is strongly dependent from each $p_i(0)$ in the initial probability distribution, not only from M, N or S.

\subsection{first-extinction time differences across theory, SDE simulation, and resampling simulation for M,N grid}

To quantify the discrepancies between theoretical predictions, the SDE simulations, and discrete resampling across a diverse set of initial conditions, we introduce a \textbf{heatmap visualization} of the differences in average first-extinction time for varying values of $M$ (number of bins) and $N$ (samples per iteration). The heatmap highlights the stability of the theoretical formula and the decreasing error with increasing $N$, confirming the reliability of our Resampling formula even for moderate system sizes.

\subsection{Scaling-like behavior for $M$, $N$ and normalized entropy $S$}
\begin{figure}
\centering
\includegraphics[width=0.7\textwidth]{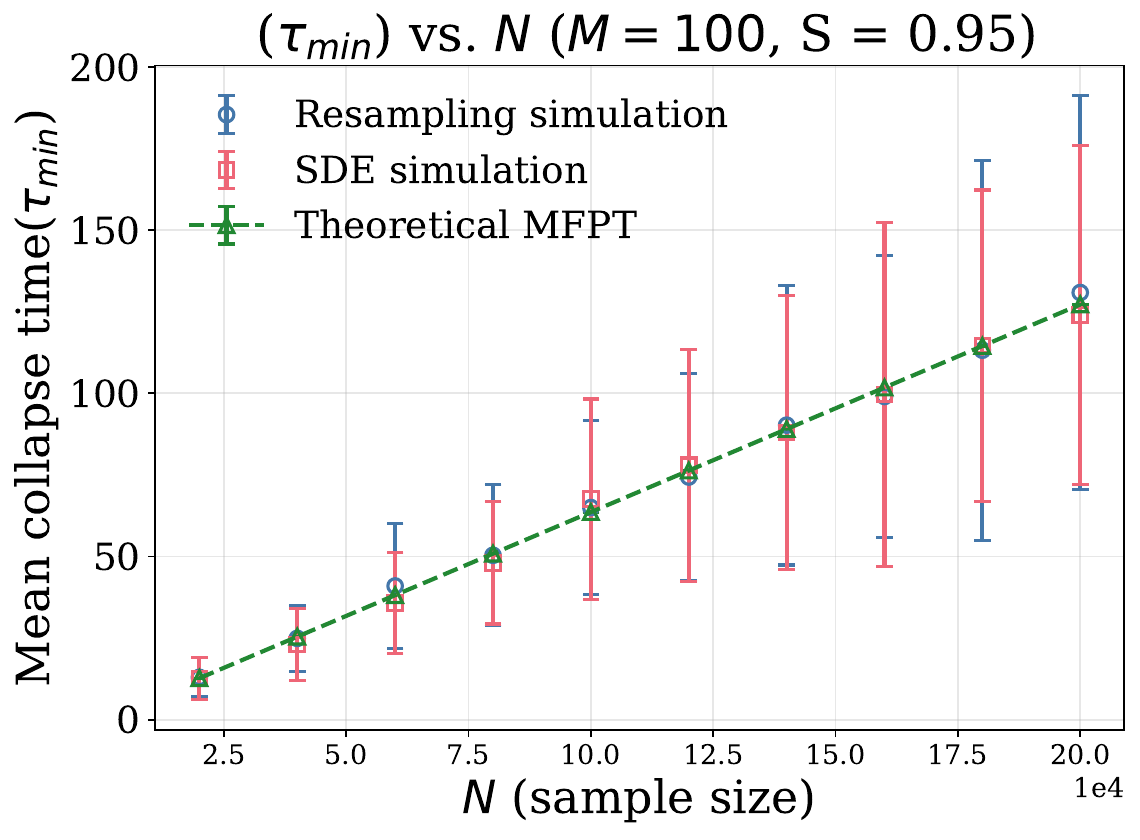}
\caption{first-extinction time as a function of number of samples N. Results are averaged over 10 independent resampling simulations for each value of $N$, averaged over 10 different initial probability distributions at initial normalized entropy $S=0.95$ .}
\label{fig:tvsN}
\end{figure}

\begin{figure}
\centering
\includegraphics[width=0.7\textwidth]{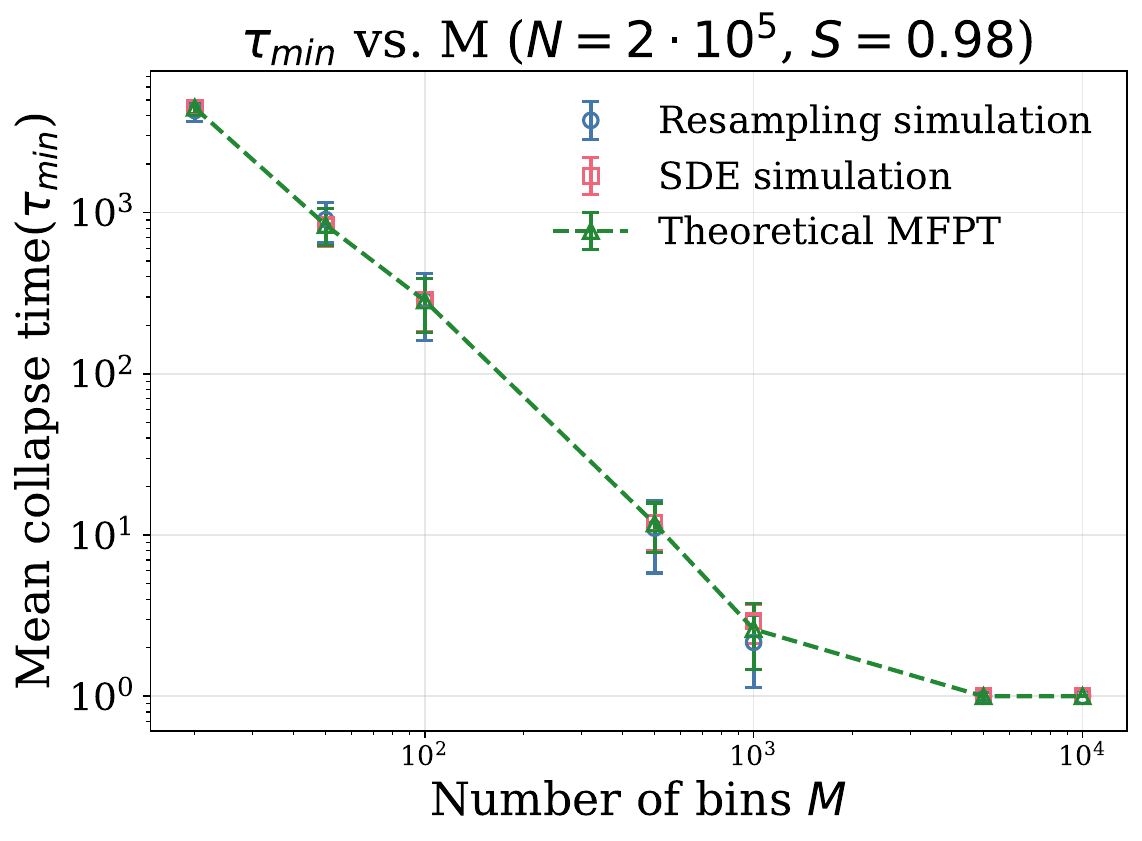}
\caption{first-extinction time as a function of number of number of states M. Results are averaged over 10 independent resampling simulations for each value of $N$, average over 10 different initial probability distributions at initial normalized entropy $S=0.98$ .}
\label{fig:tvsM}
\end{figure}

\begin{figure}
\centering
\includegraphics[width=0.7\textwidth]{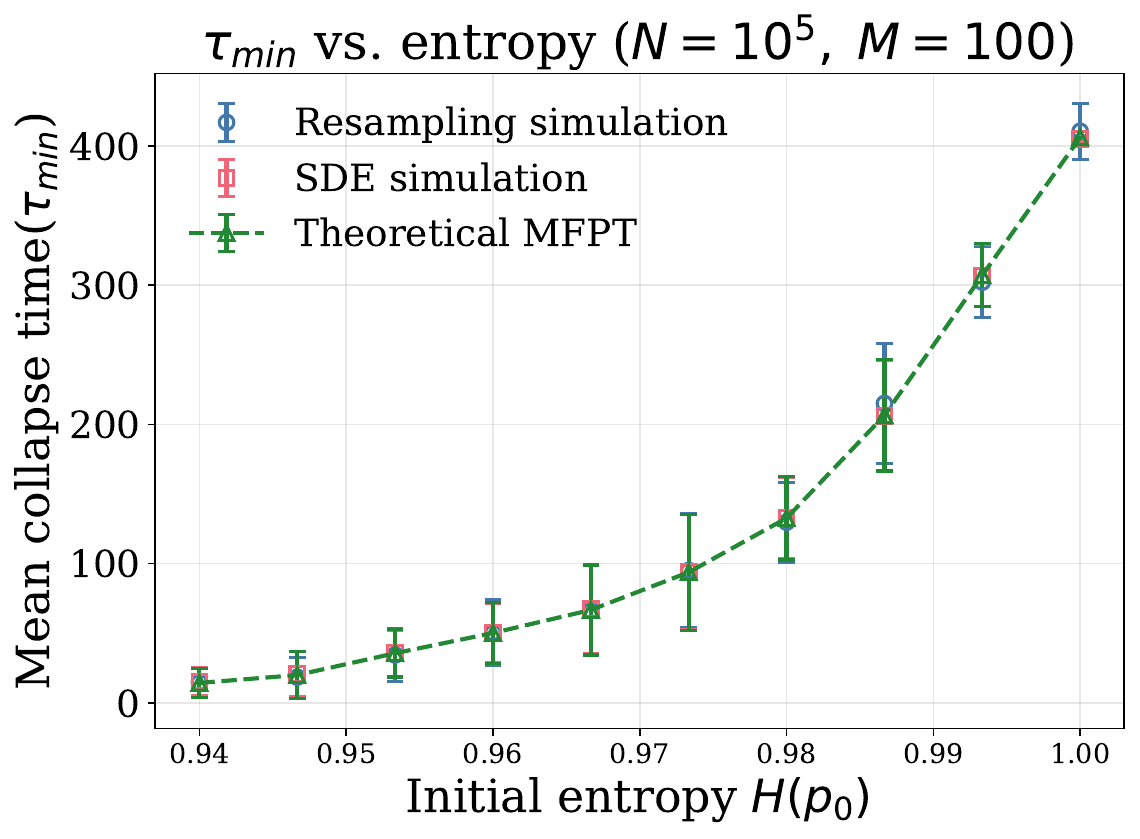}
\caption{The first-extinction time is shown as a function of the normalized entropy $S$ of the initial distribution. For each value of $S$, we average the results over 20 different initial probability distributions, and for each of these, we run 20 independent resampling simulations.}
\label{fig:tvsS}
\end{figure}

We further extended our analysis by simulating the resampling collapse as a function of \( N \)(Fig. \ref{fig:tvsN}), \( M \)(Fig. \ref{fig:tvsM}) and the normalized initial entropy $S$ (Fig. \ref{fig:tvsS}). The results confirm a clear linear scaling with \( N \), and suggest the existence of a scaling law with respect to \( M \), although its exact form remains to be determined. The same can be said for $S$. It is relevant to notice that, as for $M$ and $N$, entropy alone is not sufficient to determine the first-extinction time, and the values represented are a mean of many initial $\{p_i(0)\}$, and initial distribution with the same entropy can have very different first-extinction times.

\subsection{Small $N$ limit}
\begin{figure}[t]
\centering
\includegraphics[width=0.7\columnwidth]{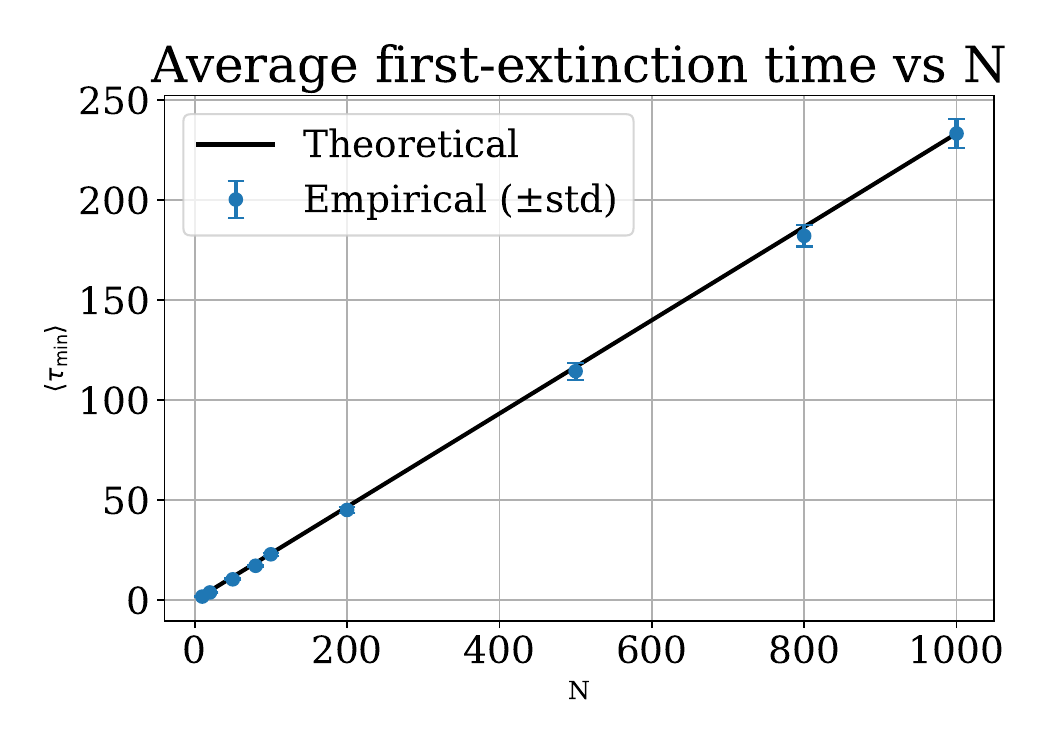}
\caption{Mean extinction time $\langle\tau_{\min}\rangle$ for flat initialization at $M=5$. 
Symbols: resampling simulations with $10^{3}$ independent realizations; solid line: theoretical prediction. 
Although the theoretical $\langle\tau_{min}\rangle$ slightly overestimates the real first-extinction time for small $N$, Excellent agreement is observed already for moderate samples sizes $N$.}
\label{fig:mean_validation}
\end{figure}

\begin{figure}[t]
\centering
\includegraphics[width=0.7\columnwidth]{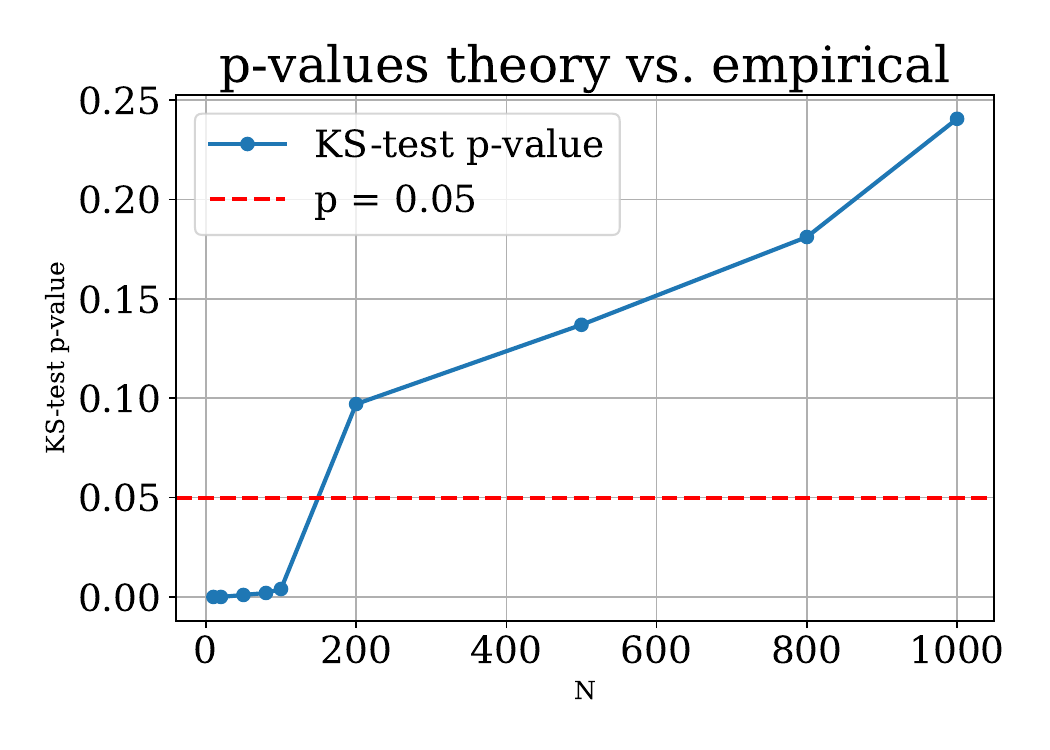}
\caption{Kolmogorov--Smirnov $p$-value for the distribution of $\tau_{\min}$ under flat initialization at $M=5$, as a function of population size $N$. 
The number of realizations used to estimate the empirical CDF is fixed at $10^{3}$. 
For $N<200$ the $p$-value falls below 0.05, indicating statistically significant deviations, while for $N\ge200$ the formula and the simulations are indistinguishable within statistical resolution.}
\label{fig:ks_pvalue}
\end{figure}

To assess the validity for small $N$, we benchmarked it against resampling simulations at flat initialization (the case with highest covariance) with $M=5$ with increasing $10\leq N \leq1000 $. Figure~\ref{fig:mean_validation} shows that the, while the theoretical predicted mean extinction time $\langle\tau_{\min}\rangle$ tends to overstimate slightly the actual first time collapse, it still agrees closely with simulations, confirming that the scaling is captured correctly. Figure~\ref{fig:ks_pvalue} reports the Kolmogorov--Smirnov $p$-value for the full distribution of $\tau_{\min}$. While for $N<200$ the $p$-value falls below 0.05, signaling statistically significant deviations, for $N\ge200$ the $p$-value rises above the threshold, indicating that the theoretical formula and the simulations become indistinguishable within statistical resolution.

\subsection{Small $M$ limit}
We also tested for which values of $M$  our formula might break down. As shown in Fig.~\ref{fig:collapse_smallM}, the resampling simulations and theory's mean first-extinction times remain in close agreement even for very small $M$, although for $M = 3,4$ the discretization effects are strong and the agreement between theoretical formula and simulations breaks down.

\subsection{Examples of CDF tests on limit configurations}
\begin{figure}[h!]
\centering
\includegraphics[width=0.7\textwidth]{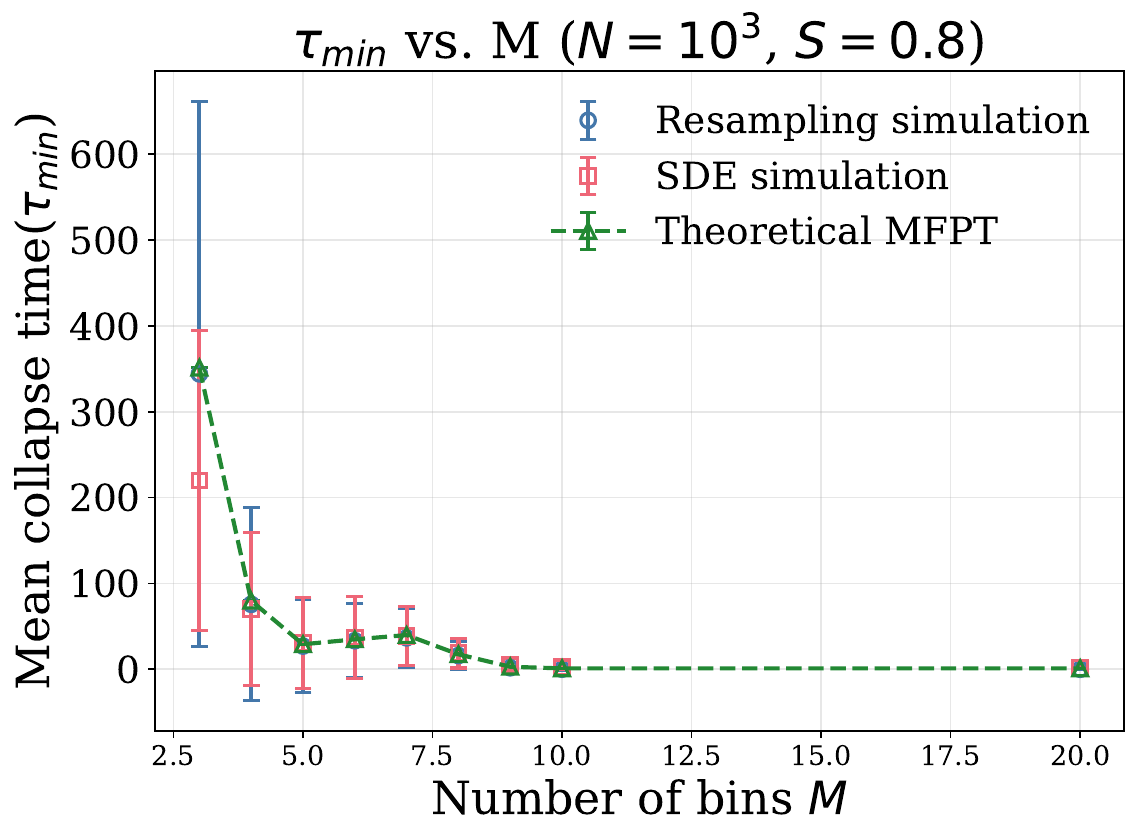}
\caption{first-extinction time as a function of the number of states $M$, for $M$ between 3 and 20, with $N = 1000$ and initial normalized entropy $S = 0.8$.}
\label{fig:collapse_smallM}
\end{figure}

\begin{figure}[ht]
\centering
\resizebox{\textwidth}{!}{%
\begin{minipage}{1.0\textwidth}

\centering
\begin{minipage}[t]{0.32\textwidth}
  \centering
  \includegraphics[width=\textwidth]{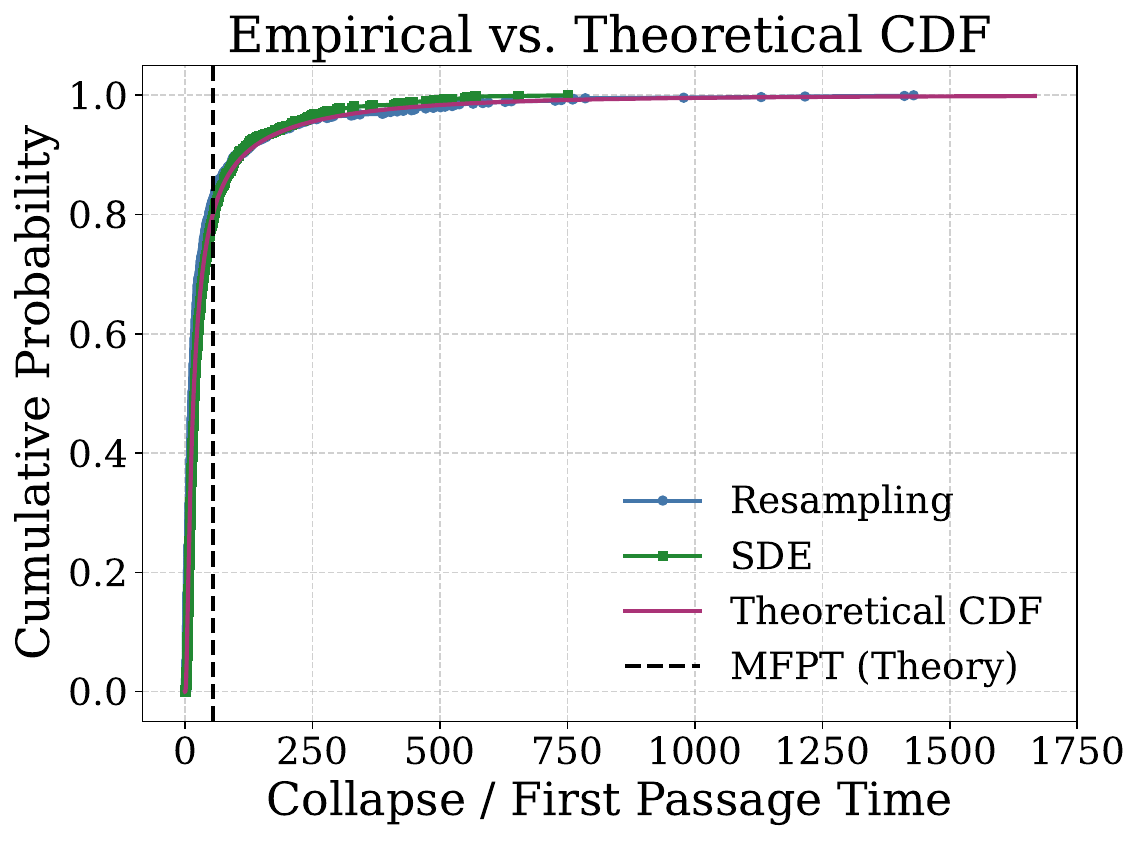}
  \vspace{0.3em}
  \small\bfseries $M=3$, $N=10^3$, $S=0.6$
\end{minipage}
\hfill
\begin{minipage}[t]{0.32\textwidth}
  \centering
  \includegraphics[width=\textwidth]{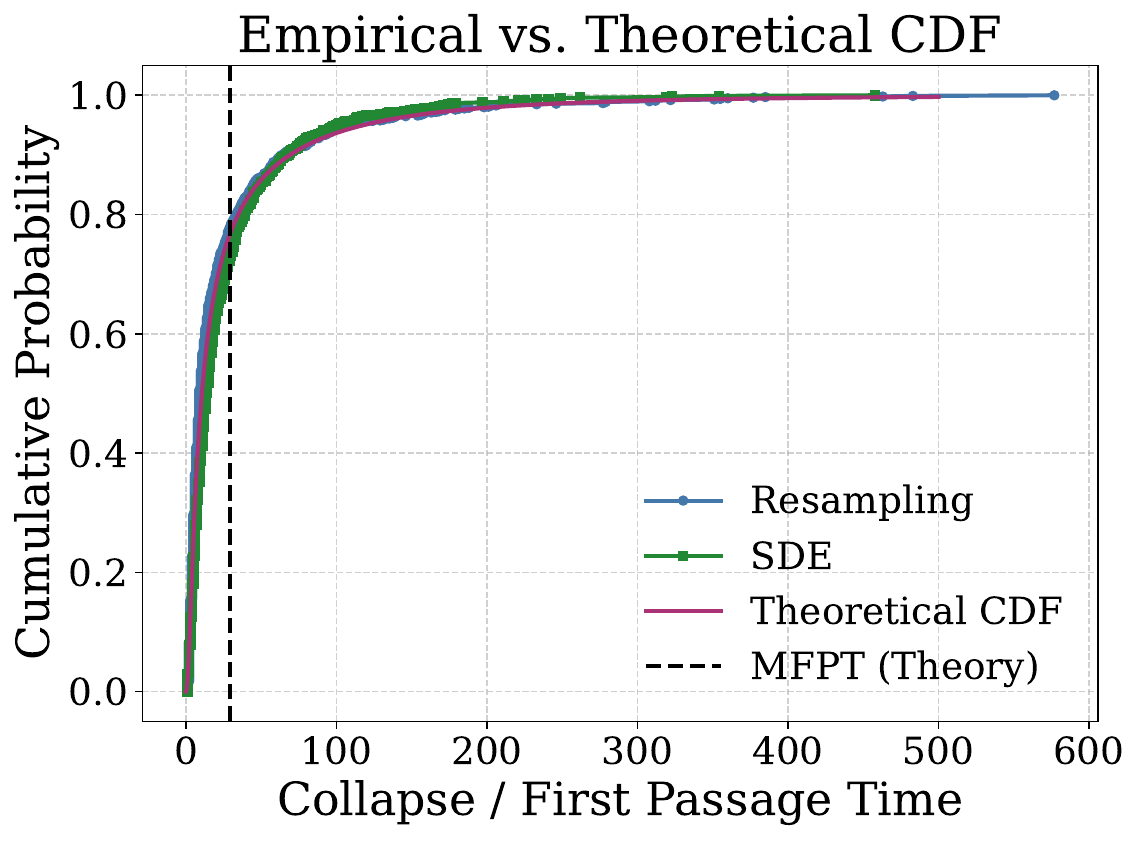}
  \vspace{0.3em}
  \small\bfseries $M=4$, $N=10^3$, $S=0.6$
\end{minipage}
\hfill
\begin{minipage}[t]{0.32\textwidth}
  \centering
  \includegraphics[width=\textwidth]{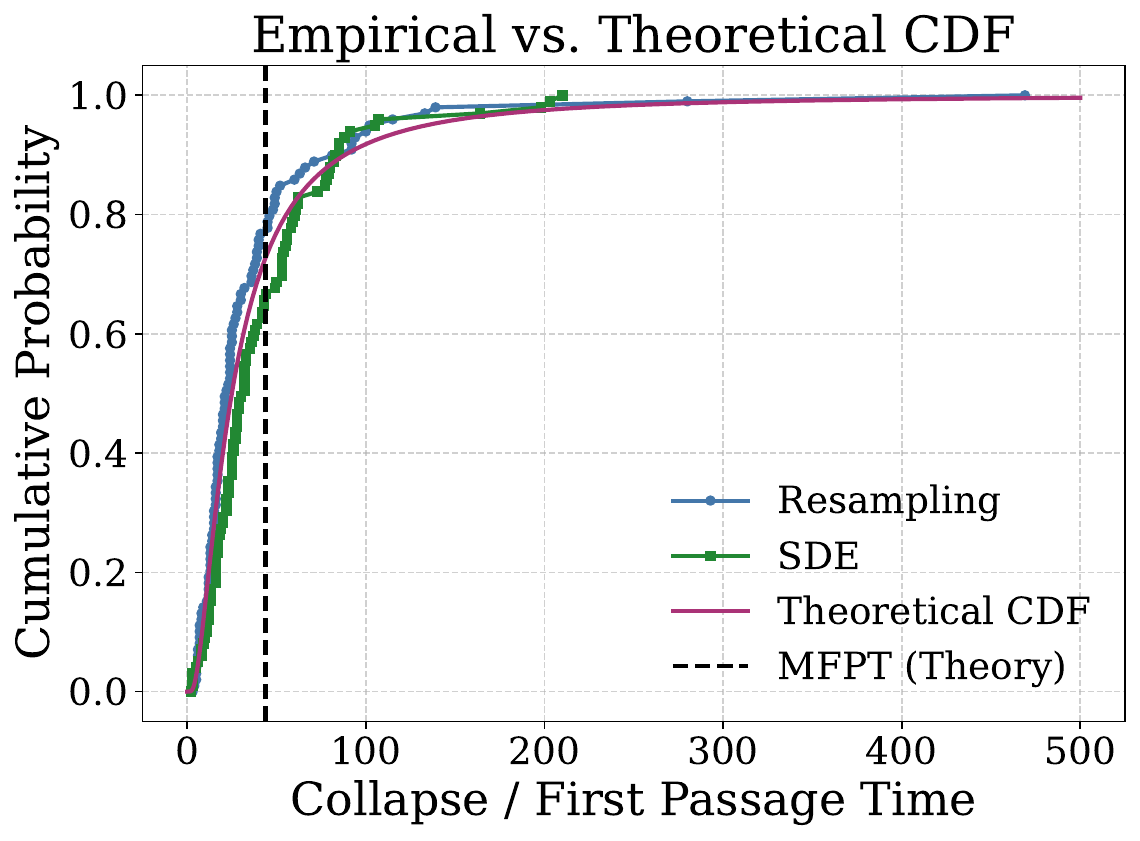}
  \vspace{0.3em}
  \small\bfseries $M=5$, $N=5\cdot10^3$, $S=0.6$
\end{minipage}

\vspace{1.5em}

\centering
\begin{minipage}[t]{0.32\textwidth}
  \centering
  \includegraphics[width=\textwidth]{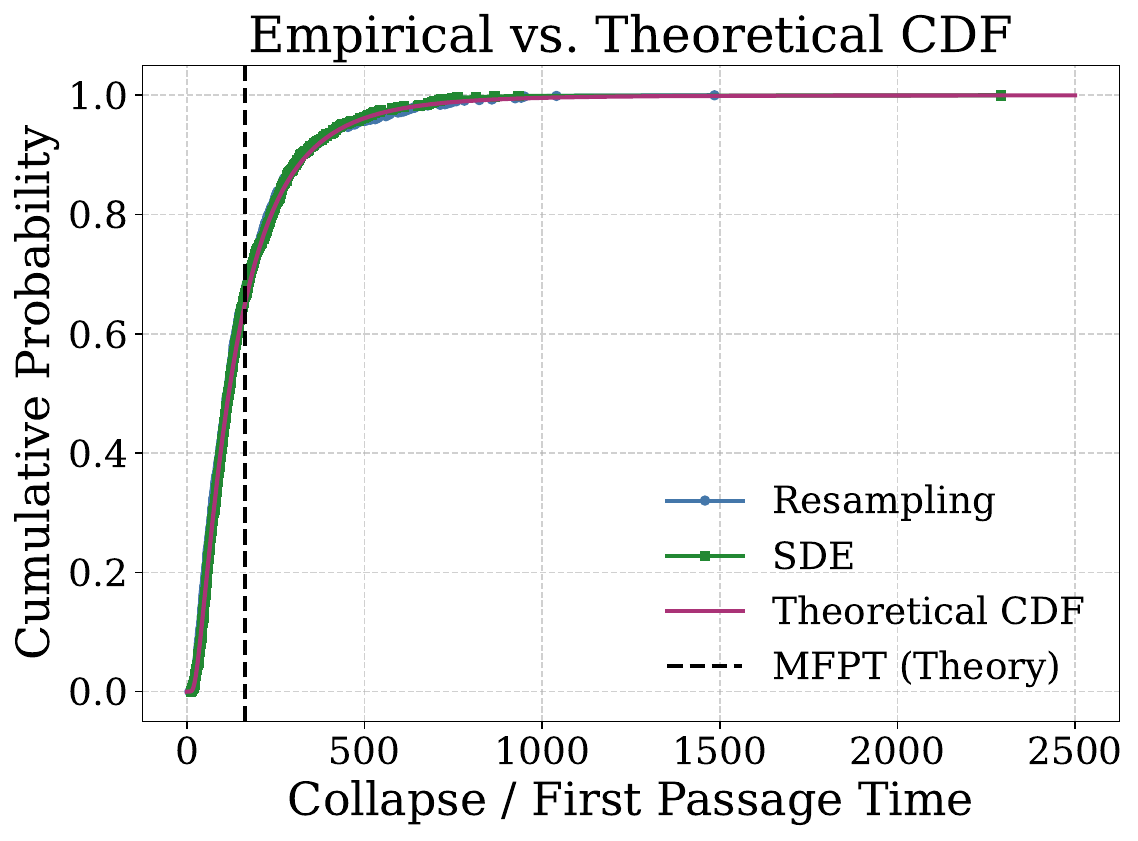}
  \vspace{0.3em}
  \small\bfseries $M=6$, $N=5\cdot10^3$, $S=0.6$
\end{minipage}
\hfill
\begin{minipage}[t]{0.32\textwidth}
  \centering
  \includegraphics[width=\textwidth]{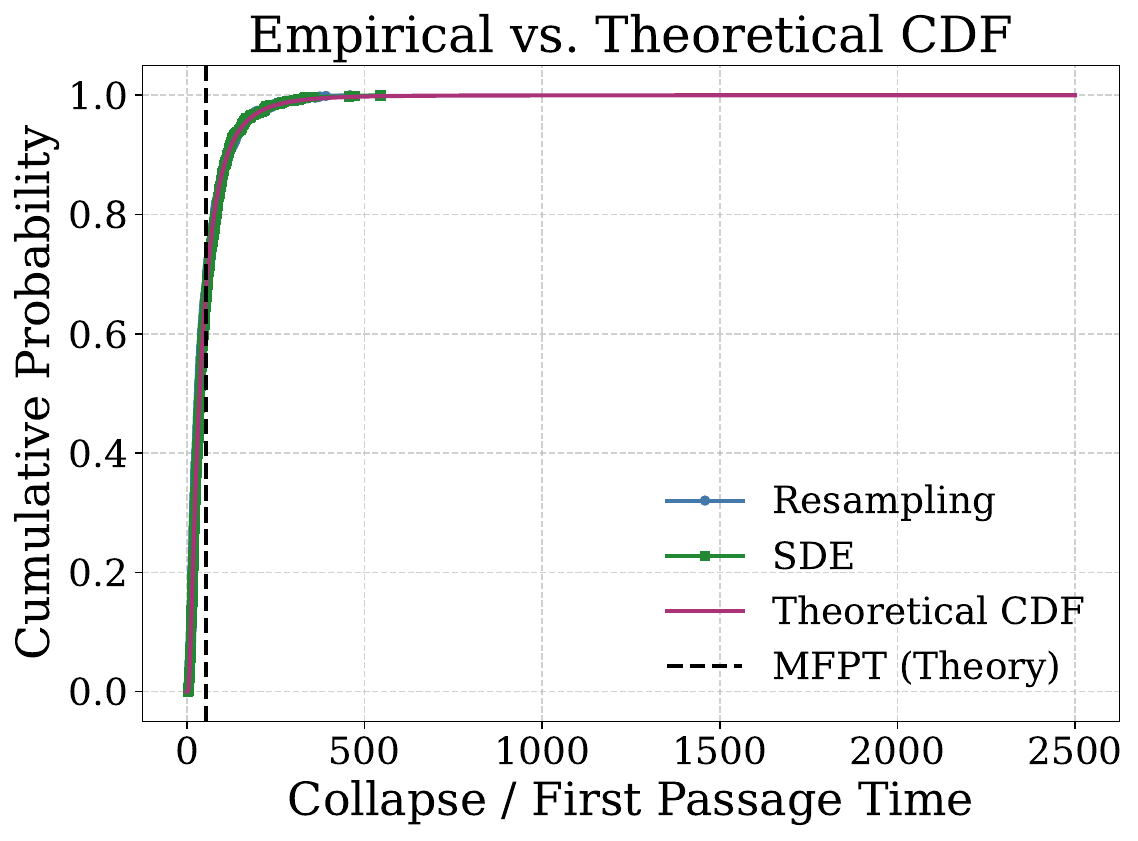}
  \vspace{0.3em}
  \small\bfseries $M=8$, $N=10^4$, $S=0.6$
\end{minipage}
\hfill
\begin{minipage}[t]{0.32\textwidth}
  \centering
  \includegraphics[width=\textwidth]{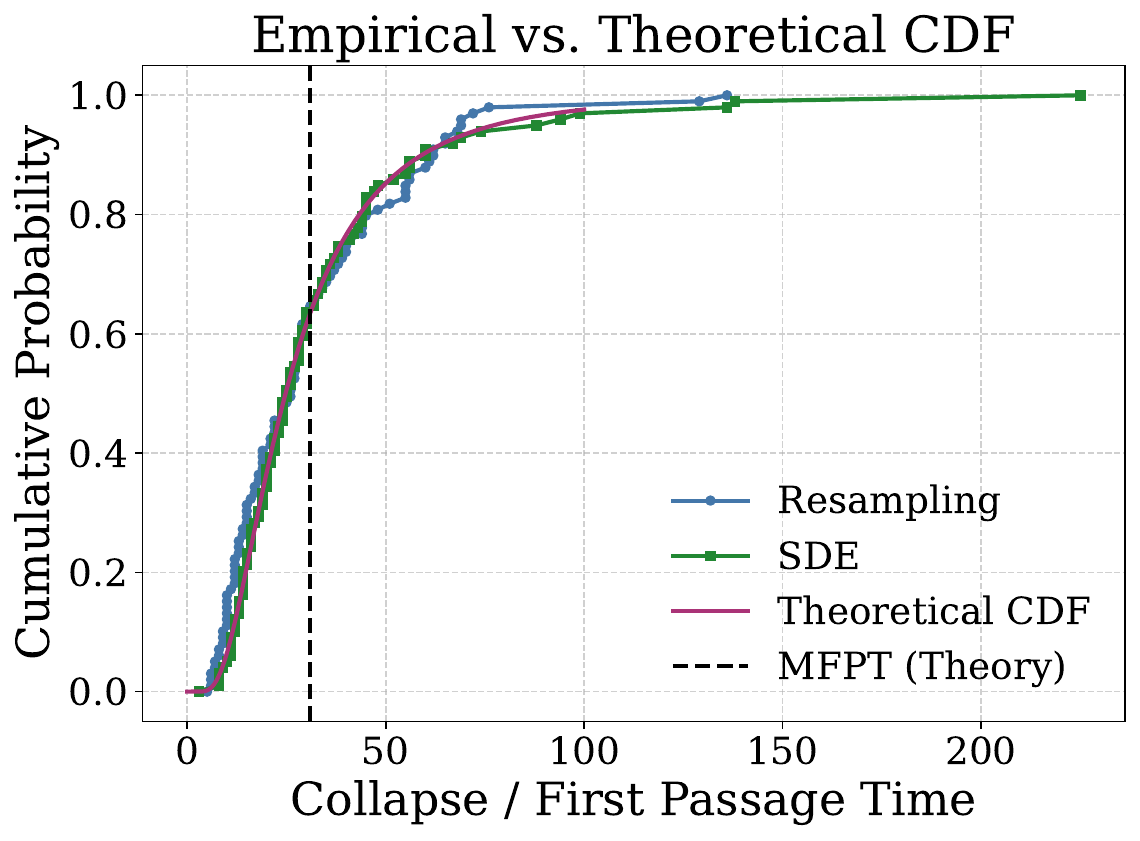}
  \vspace{0.3em}
  \small\bfseries $M=10$, $N=10^4$, $S=0.7$
\end{minipage}

\vspace{1.5em}

\centering
\begin{minipage}[t]{0.32\textwidth}
  \centering
  \includegraphics[width=\textwidth]{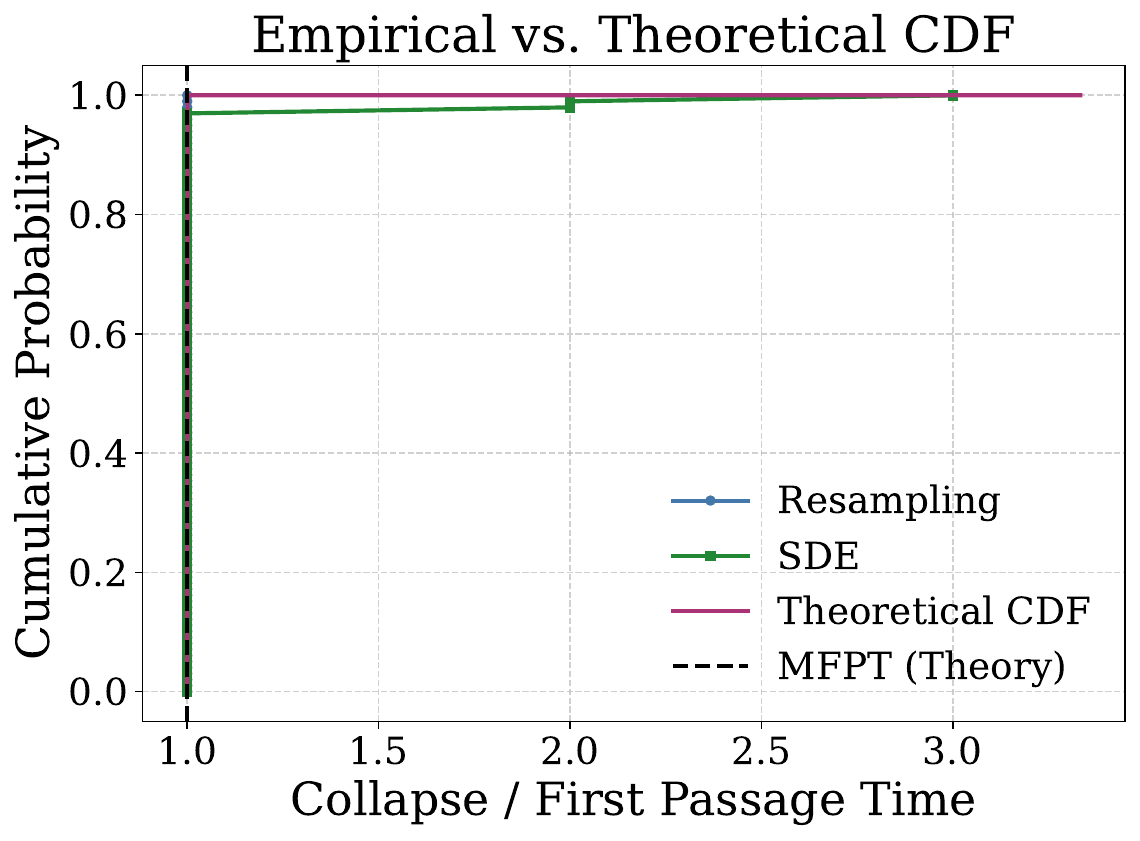}
  \vspace{0.3em}
  \small\bfseries $M=10$, $N=5\cdot10^6$, $S=0.3$
\end{minipage}
\hfill
\begin{minipage}[t]{0.32\textwidth}
  \centering
  \includegraphics[width=\textwidth]{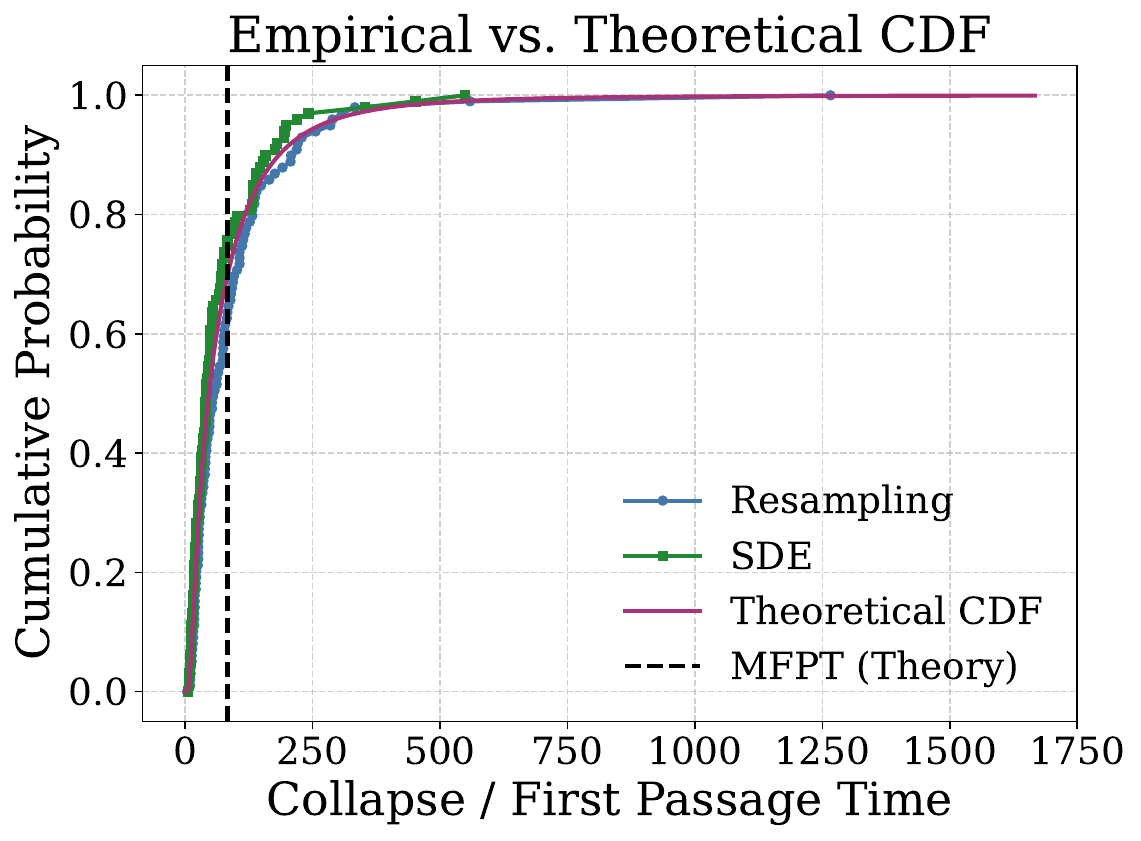}
  \vspace{0.3em}
  \small\bfseries $M=10$, $N=10^7$, $S=0.3$
\end{minipage}
\hfill
\begin{minipage}[t]{0.32\textwidth}
  \centering
  \includegraphics[width=\textwidth]{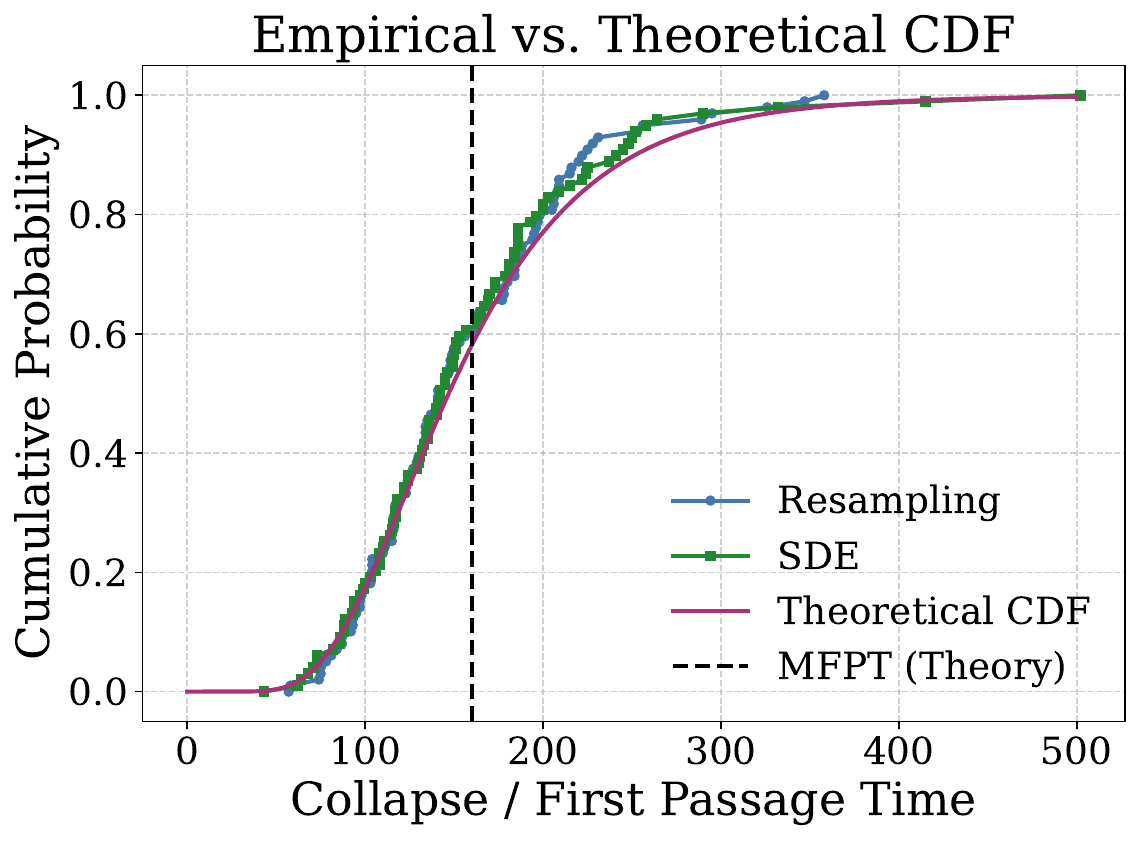}
  \vspace{0.3em}
  \small\bfseries $M=10$, $N=2\cdot10^3$, $S=0.999$
\end{minipage}

\vspace{1.5em}

\centering
\begin{minipage}[t]{0.32\textwidth}
  \centering
  \includegraphics[width=\textwidth]{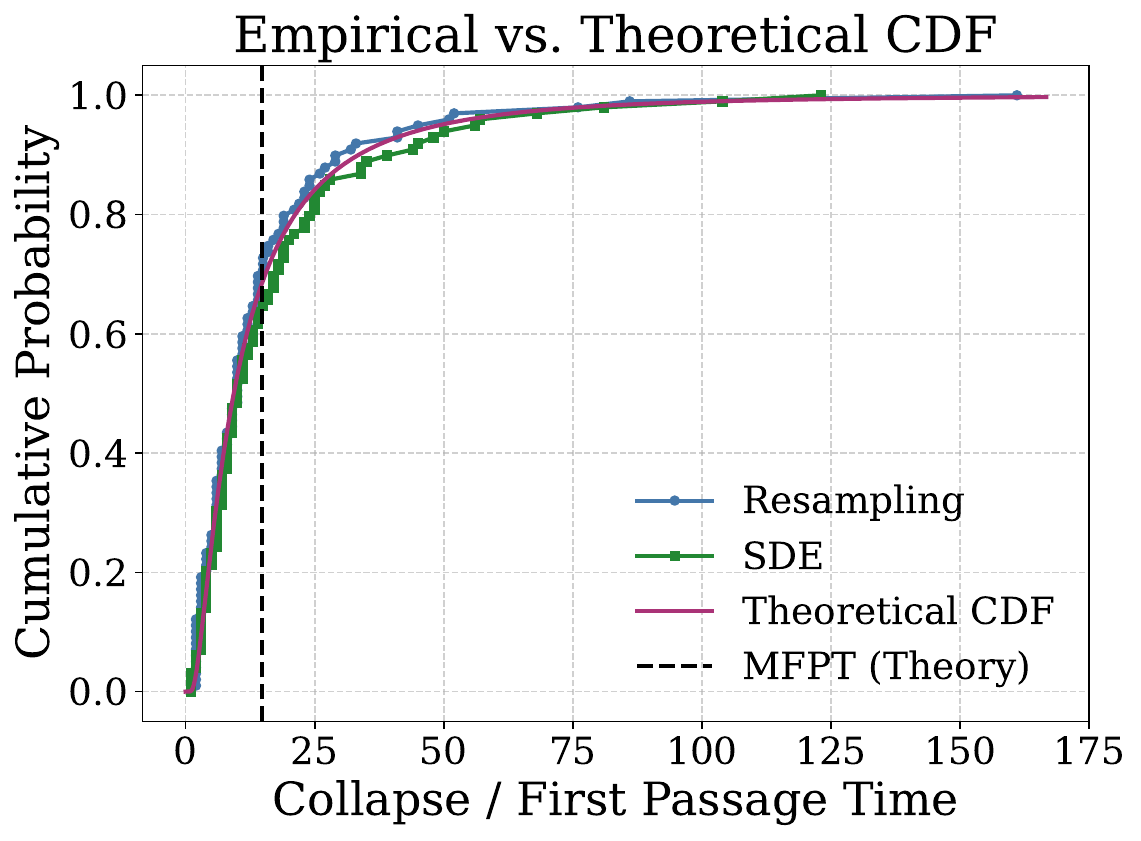}
  \vspace{0.3em}
  \small\bfseries $M=25$, $N=10^6$, $S=0.8$
\end{minipage}
\hfill
\begin{minipage}[t]{0.32\textwidth}
  \centering
  \includegraphics[width=\textwidth]{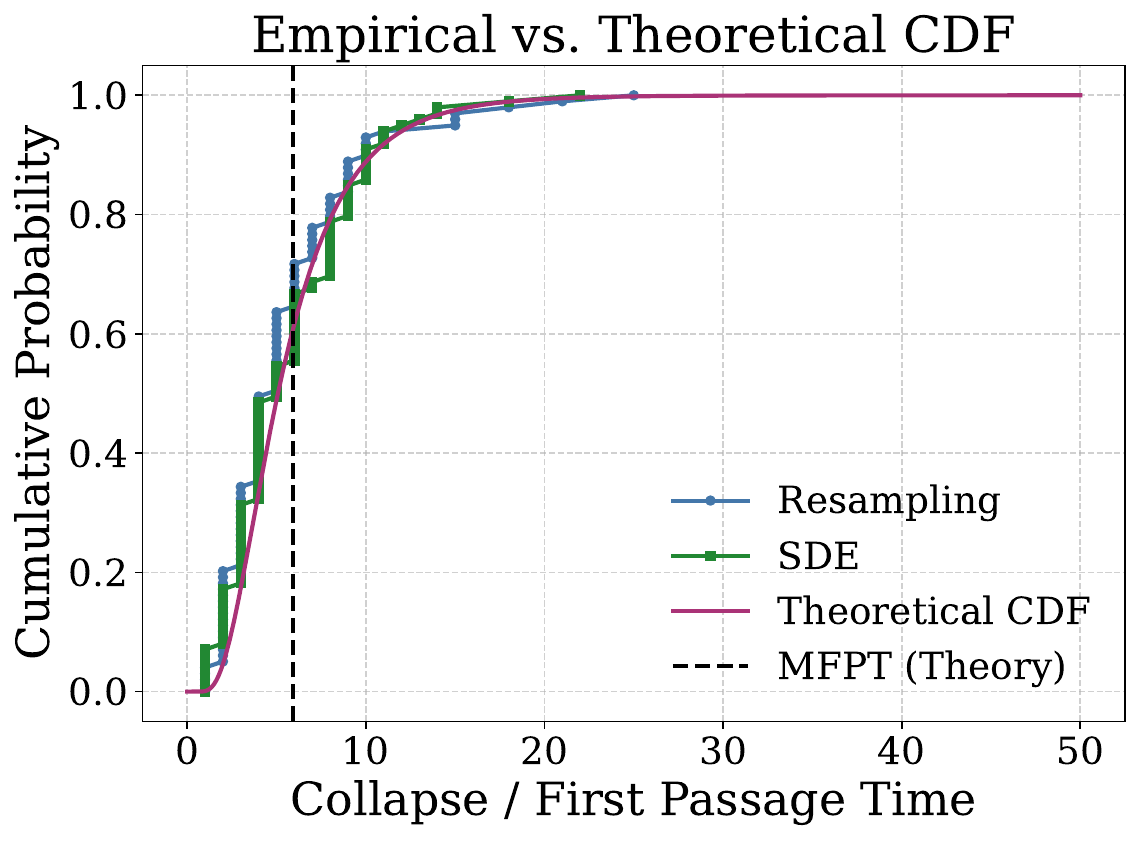}
  \vspace{0.3em}
  \small\bfseries $M=100$, $N=5\cdot10^7$, $S=0.8$
\end{minipage}
\hfill
\begin{minipage}[t]{0.32\textwidth}
  \centering
  \includegraphics[width=\textwidth]{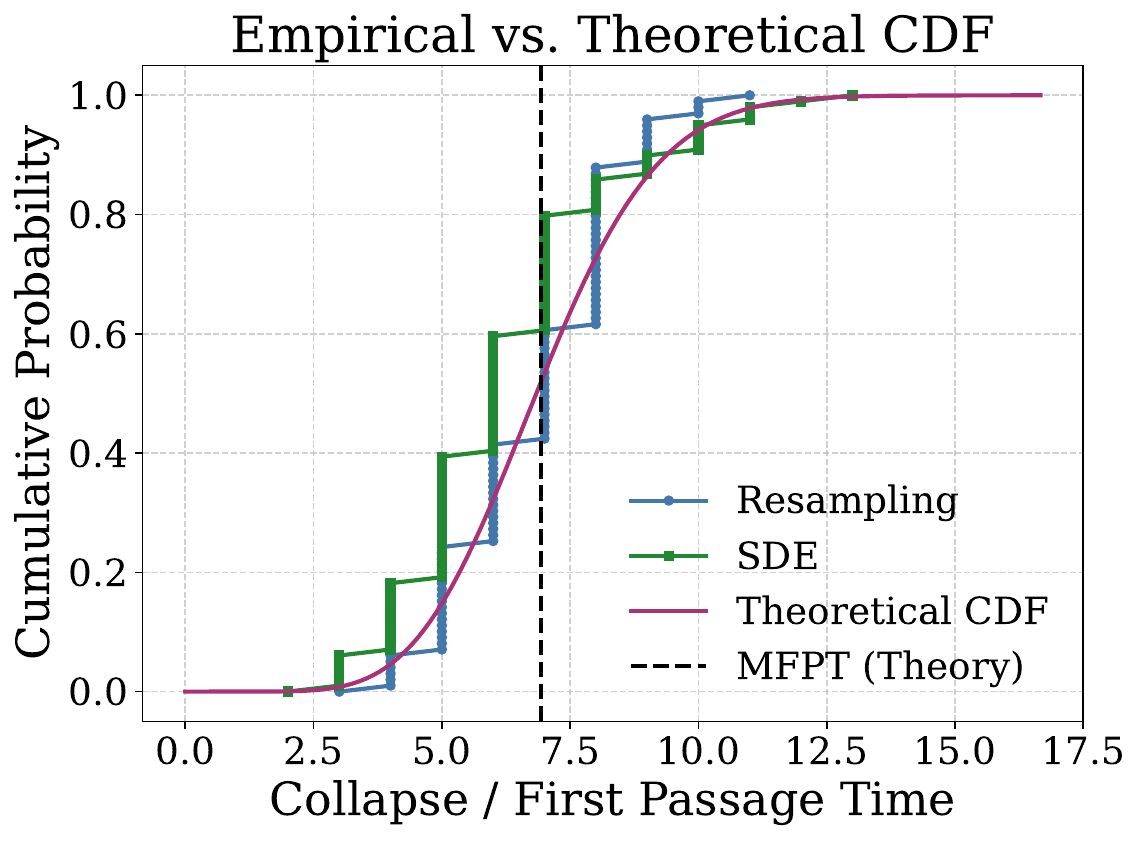}
  \vspace{0.3em}
  \small\bfseries $M=100$, $N=2\cdot10^3$, $S=0.999$
\end{minipage}

\end{minipage}
} 

\caption{Empirical vs.\ theoretical cumulative distribution functions (CDFs) for first-extinction times (first-extinction times = collapse/first passage time) under various combinations of $M$, $N$, and $S$. Labels below each plot indicate the parameter values used in each simulation.}
\label{fig:multi_cdf}
\end{figure}

We further tested the predictive capabilities of our theoretical framework by deliberately selecting limiting configurations for the resampling experiments, including very small values of $M$, diverse initial entropies, and limited sample sizes. As shown in Fig.~\ref{fig:multi_cdf} and Table ~\ref{tab:simres}, the results remain statistically robust and the theoretical predictions are in good agreement with the resampling simulations in nearly all cases. Exceptions occur only for $M = 3$ and $M = 4$, where the Kolmogorov–Smirnov test indicates a slight discrepancy: the p-value is below 0.05 for $M = 3$ and marginally above 0.05 for $M = 4$. In these specific cases, the mean first-extinction times from the resampling differ from the theoretical estimate by approximately $9\%$. This explains the apparent contradiction of Fig.~\ref{fig:collapse_smallM}: while the mean first-extinction time might be compatible between theory and actual resampling even for small $M$, the shape the actual CDFs are not similar enough to get a $p\geq0.05$, thus making them not compatible. For $M \geq 5$, the simulations are in excellent agreement with the theory for a wide range of values for $S$ and $N$, both in terms of mean values and full distributional shape, as confirmed by the KS test applied to the empirical cumulative distribution functions (CDFs).

\section{Network Architecture}

The predictive model used to validate the resampling collapse theory is a feed-forward neural network with the following configuration:

\begin{itemize}
    \item \textbf{Architecture:} Fully connected feed-forward network with two hidden layers, each consisting of 50 neurons.
    \item \textbf{Input/Output dimension:} 30 nodes (corresponding to the number of discrete states in the Markov chain).
    \item \textbf{Activation functions:} ReLU for hidden layers; softmax at output to obtain a probability vector.
    \item \textbf{Optimizer:} Adam optimizer with a learning rate of 0.01.
    \item \textbf{Training configuration:} Each training chunk consisted of 1000 samples, and the network was trained for 300 epochs.
    \item \textbf{Initial entropy:} The initial normalized entropy of the prediction distribution was 0.85.
    \item \textbf{Experimental runs:} The resampling collapse experiments were conducted with sample sizes of  20{,}000, 50{,}000 and 100{,}000 respectively.
\end{itemize}

\begin{figure}[ht]
\centering
\includegraphics[width=0.48\textwidth]{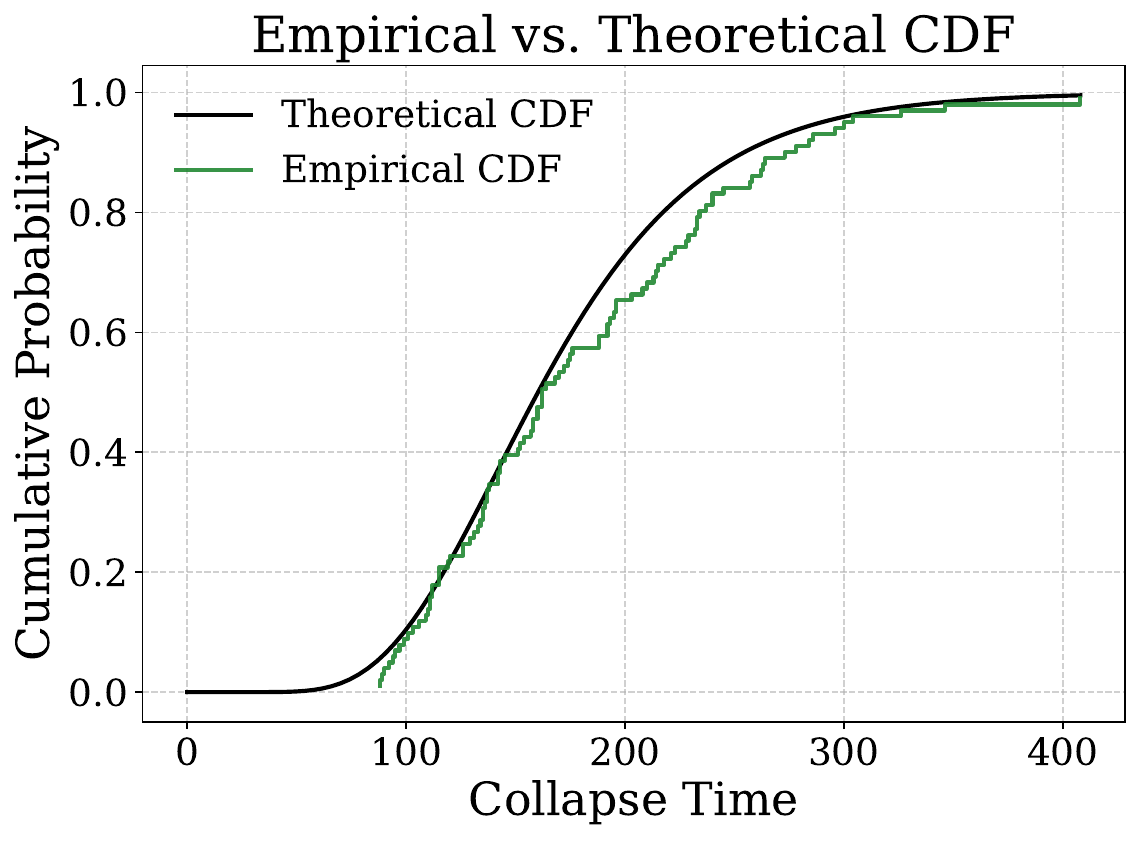}
\hfill
\includegraphics[width=0.48\textwidth]{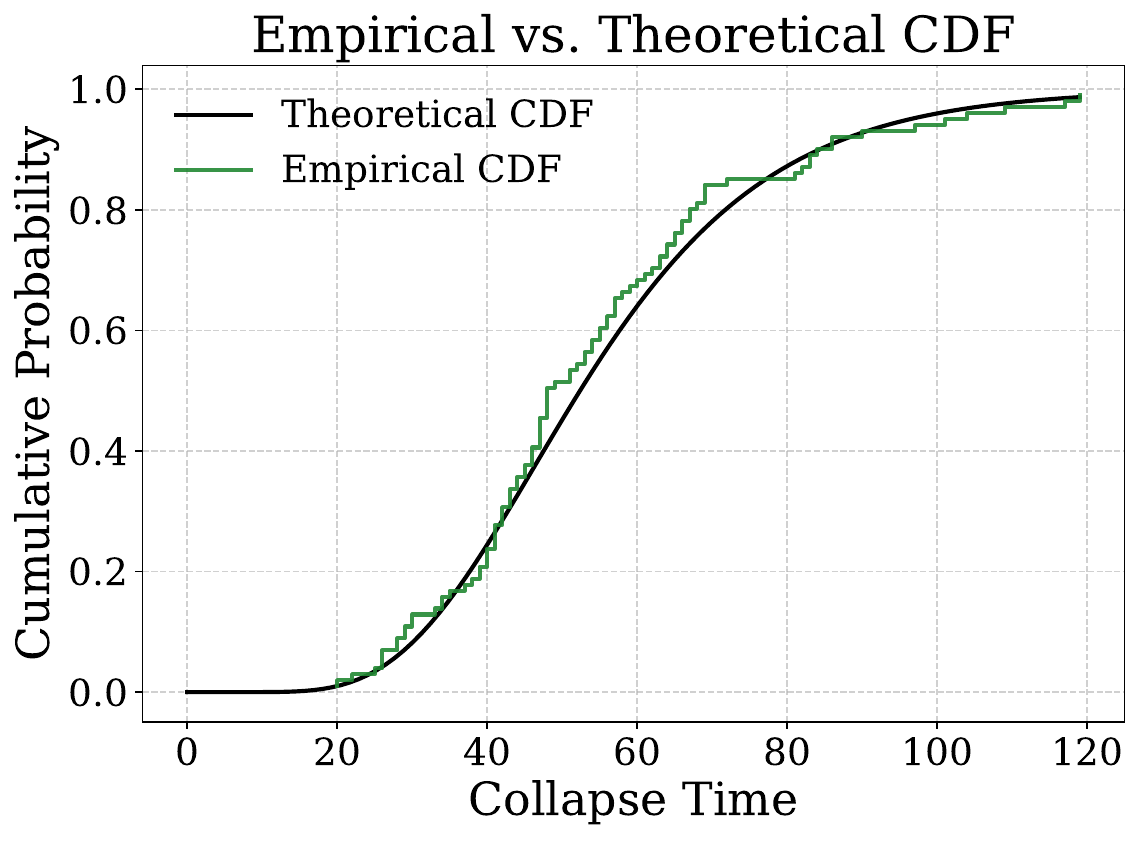}
\caption{
Comparison between the empirical cumulative distribution functions (CDFs) of the first-extinction times obtained from the Markov AI resampling simulations (green) and the theoretical predictions (black) for the limit configurations with $M = 30$ and $S = 0.85$. 
Left: $N = 20{,}000$. 
Right: $N = 50{,}000$. 
}
\label{fig:markov_ai_cdf_M30}
\end{figure}

As pointed out in the paper, the total number of trainings used in the $N=100{,}000$ setting was $ 31{,}054$ for one CDF. The big number of trainings necessary to get a statistically significant sample for model collapse explains the constraint of our analysis to relatively small architectures. Still, this is sufficient to demonstrate the progressive loss of low-frequency states due to iterative prediction and sampling, consistent with the theoretical formula derived in the main text. As shown in Fig.~\ref{fig:markov_ai_cdf_M30}, two more experiments have been conducted, for $N=2\cdot10^4$ and $N=5\cdot10^4$, to further solidify our claims. The KS test gave $KS_{statistic}(N=2\cdot10^4)=0.10$ and $KS_{statistic}(N=5\cdot10^4)=0.09$, and $p_{value}(N=2\cdot10^4)=0.702$ and $p_{value}(N=5\cdot10^4)=0.815$. Code and data will be available upon reasonable request to the authors.

\begin{figure}
\centering
\includegraphics[width=0.45\textwidth]{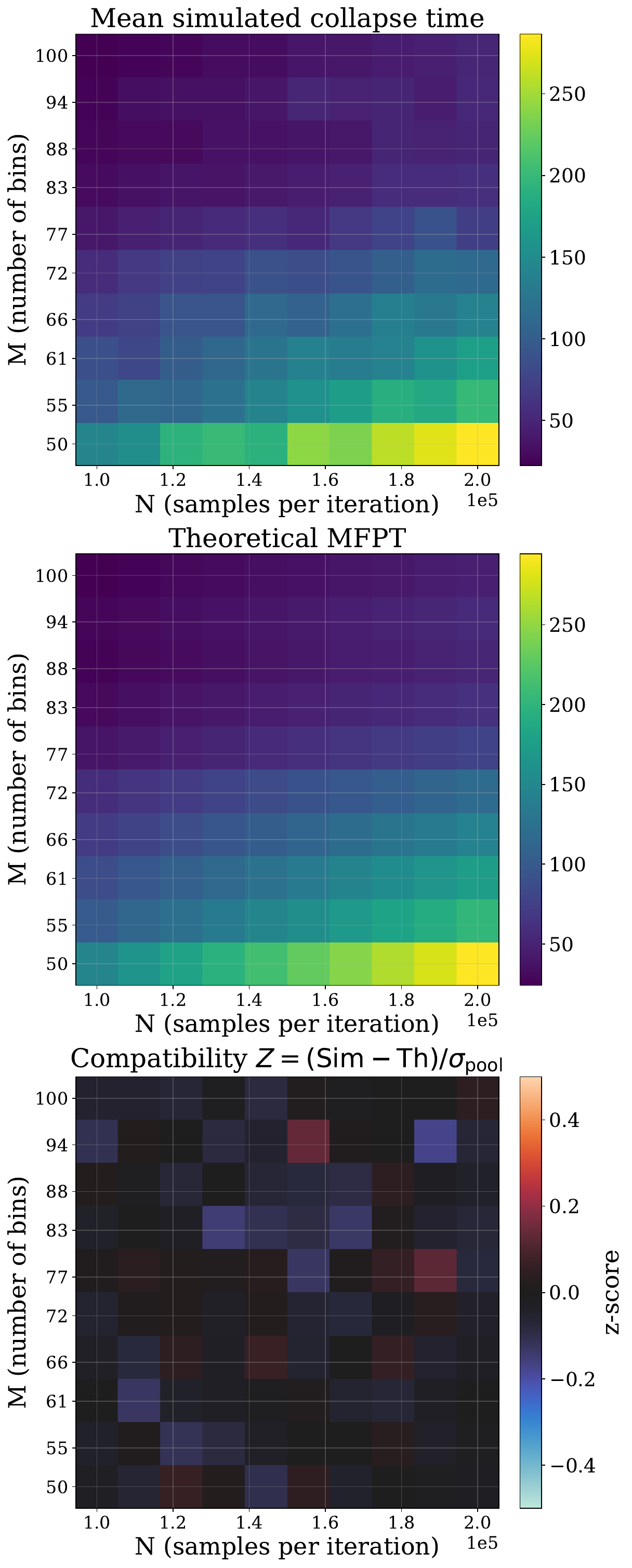}
\caption{\textbf{first-extinction time analysis across $(M, N)$:} Mean simulated first-extinction time (top), theoretical prediction (middle), and $z$-compatibility score $(\langle \tau_{\text{sim}}\rangle - \langle\tau_{theory }\rangle)/\sigma_{\text{pool}}$ (bottom). Each cell shows the average over 10 different initial probability distributions (all with normalized entropy $S = 0.95$), each simulated 10 times. for z-scores $\leq 1$ Theory and simulations are compatible.}
\label{fig:heatmpap}
\end{figure}


\begin{table}[ht]
\centering
\caption{Summary of simulation results across 12 configurations. The results for $M\leq10$ are all on $10^3$ simulations, while the others are on $10^2$ simulations.}
\resizebox{1.0\textwidth}{!}{
\begin{tabular}{|l|c|c|c|}
\toprule
\textbf{Metric} & \textbf{M=3, N=$10^3$, S=0.6} & \textbf{M=4, N=$10^3$, S=0.6} & \textbf{M=5, N=$5\cdot10^3$, S=0.6} \\
\midrule
Mean Resampling & 51.66 & 27.6& 38.03\\
Std. Resampling & 4.04 & 1.71& 5.78\\
Mean SDE & 46.65& 27.82& 42.82\\
Std. SDE & 4.363 & 1.42& 3.92\\
Theor. $\langle \tau_{\min} \rangle$ & 56.30 & 30.26& 43.21\\
KS Stat SDE vs Resamp & 0.094 & 0.1270& 0.22\\
KS P-Val SDE vs Resamp & 0.003 & 0.0000& 0.0156\\
KS Stat Theor vs SDE & 0.0830 & 0.0780& 0.15\\
KS P-Val Theor vs SDE & 0.0156& 0.0045& 0.212\\
KS Stat Resamp vs Theor & 0.0610 & 0.0600& 0.090\\
KS P-Val Resamp vs Theor & 0.0484 & 0.0546& 0.8154\\
\midrule
\textbf{Metric} & \textbf{M=6, N=$5\cdot10^3$, S=0.6}& \textbf{M=8, N=$5\cdot10^3$, S=0.6}& \textbf{M=10, N=$10^4$, S=0.7} \\
\midrule
Mean Resampling & 160.33& 53.06& 30.82 \\
Std. Resampling & 4.91& 1.75& 2.3 \\
Mean SDE & 160.70& 54.53& 33.56 \\
Std. SDE & 4.84& 1.78& 3.07 \\
Theor. $\langle \tau_{\min} \rangle$ & 165.20& 54.73& 30.89 \\
KS Stat SDE vs Resamp & 0.0290& 0.0370& 0.11 \\
KS P-Val SDE vs Resamp & 0.7947& 0.5006& 0.583 \\
KS Stat Theor vs SDE & 0.0300& 0.0280& 0.09 \\
KS P-Val Theor vs SDE & 0.7594& 0.8282& 0.8154 \\
KS Stat Resamp vs Theor & 0.0230& 0.0270& 0.14 \\
KS P-Val Resamp vs Theor & 0.9542& 0.8281& 0.2819 \\
\midrule
\textbf{Metric} & \textbf{M=10, N=$5\cdot10^6$, S=0.3} & \textbf{M=10, N=$10^7$, S=0.3}& \textbf{M=10, N=$2\cdot10^3$, S=0.999} \\
\midrule
Mean Resampling & 1 & 95.15 & 154.64 \\
Std. Resampling & 0 & 14.6 & 6.014 \\
Mean SDE & 1.04 & 87.14 & 156.27 \\
Std. SDE & 0.024 & 8.693 & 7.060 \\
Theor. $\langle \tau_{\min} \rangle$ & 1 & 93.12 & 161.08 \\
KS Stat SDE vs Resamp & 0.03 & 0.17 & 0.05 \\
KS P-Val SDE vs Resamp & 1 & 0.112 & 0.9992 \\
KS Stat Theor vs SDE & 0.02 & 0.07 & 0.09 \\
KS P-Val Theor vs SDE & 1 & 0.9684 & 0.8154 \\
KS Stat Resamp vs Theor & 0.02 & 0.11 & 0.11 \\
KS P-Val Resamp vs Theor & 1 & 0.583 & 0.581\\
\midrule
\textbf{Metric} & \textbf{M=25, N=$10^6$, S=0.8} & \textbf{M=100, N=$5\cdot10^7$, S=0.8} & \textbf{M=100, N=$2\cdot10^3$, S=0.999} \\
\midrule
Mean Resampling & 15.17 & 5.53 & 6.81\\
Std. Resampling & 2.058 & 0.42 & 0.160\\
Mean SDE & 16.86 & 5.74 & 6.25\\
Std. SDE & 2.04 & 0.38 & 0.213\\
Theor. $\langle \tau_{\min} \rangle$ & 15.77 & 5.92 & 7.84\\
KS Stat SDE vs Resamp & 0.08 & 0.09 & 0.19\\
KS P-Val SDE vs Resamp & 0.9084 & 0.8154 & 0.0539\\
KS Stat Theor vs SDE & 0.1 & 0.16 & 0.22\\
KS P-Val Theor vs SDE & 0.7021 & 0.1548 & 0.008\\
KS Stat Resamp vs Theor & 0.17 & 0.18 & 0.16\\
KS P-Val Resamp vs Theor & 0.1112 & 0.0782 & 0.1548\\
\bottomrule
\end{tabular}
}
\label{tab:simres}
\end{table}

\end{document}